\documentclass[lettersize,journal]{IEEEtran}
\usepackage{amsmath,amsfonts}
\usepackage{algorithmic}
\usepackage{algorithm}
\usepackage{array}
\usepackage[caption=false,font=normalsize,labelfont=sf,textfont=sf]{subfig}
\usepackage{textcomp}
\usepackage{stfloats}
\usepackage{url}
\usepackage{verbatim}
\usepackage{graphicx}
\usepackage{cite}
\usepackage{color}
\usepackage{pdfpages}

\usepackage{subfloat}

\usepackage[T1]{fontenc}% 可选T1字体编码

\usepackage{amsmath}
\usepackage[cmintegrals]{newtxmath}
\usepackage{color}

\usepackage{multirow}

\begin{document}

\title{Decision-making of Emergent Incident based on P-MADDPG\\ }

\author{Yibo Guo, Lishuo Hou, Mingxin Li, Yue Yuan, Shun Liu, Jingyi Xue, Yafang Han, Mingliang Xu\\ \emph{School of Computer and Artificial Intelligence},
\\\emph{Zhengzhou University} 

        % <-this % stops a space
\thanks{Yibo Guo, Lishuo Hou, Mingxin Li, Yue Yuan, Shun Liu, Jingyi Xue, Yafang Han, Mingliang Xu.Decision-making of Emergent Incident based on P-MADDPG. arxiv:submit/4215882, 2022}
\thanks{This paper was produced by NSFC 61602421}
\thanks{This paper was produced by China Postdoctoral Science Foundation 2016M600584.}% <-this % stops a space
%\thanks{Manuscript received April 19, 2022; revised August 16, 2021.}

}

% The paper headers
%\markboth{Journal of \LaTeX\ Class Files,~Vol.~14, No.~8, August~2021}%
%{Shell \MakeLowercase{\textit{et al.}}: A Sample Article Using IEEEtran.cls for IEEE Journals}

%\IEEEpubid{0000--0000/00\$00.00~\copyright~2021 IEEE}
% Remember, if you use this you must call \IEEEpubidadjcol in the second
% column for its text to clear the IEEEpubid mark.

\maketitle

\begin{abstract}
In recent years, human casualties and damage to resources caused by emergent incidents have become a serious problem worldwide. In this paper, we model the emergency decision-making problem and use Multi-agent System (MAS) to solve the problem that the decision speed cannot keep up with the spreading speed.
MAS can play an important role in the automated execution of these tasks to reduce mission completion time.
In this paper, we propose a P-MADDPG algorithm to solve the emergency decision-making problem of emergent incidents, which predicts the nodes where an incident may occur in the next time by GRU model and makes decisions before the incident occurs, thus solving the problem that the decision speed cannot keep up with the spreading speed.
A simulation environment was established for realistic scenarios, and three scenarios were selected to test the performance of P-MADDPG in emergency decision-making problems for emergent incidents: unmanned storage, factory assembly line, and civil airport baggage transportation.
Simulation results using the P-MADDPG algorithm are compared with the greedy algorithm and the MADDPG algorithm, and the final experimental results show that the P-MADDPG algorithm converges faster and better than the other algorithms in scenarios of different sizes.
This shows that the P-MADDP algorithm is effective for emergency decision-making in emergent incident.
\end{abstract}


\section{Introduction}

\IEEEPARstart{P}{rincipled} decision making problem in emergency response management necessitates the use of statistical models that predict the spatial-temporal likelihood of incident occurrence due to the influence of complex environment, high uncertainty, lack of information and other factors. In addition to the above difficulties, the actual dynamic scheduling is running in the stochastic environment of inevitable interruption or unpredictable events. These uncertain disturbances may disturb the pre- established optimal scheduling, or even make it infeasible. The original task sequence is affected when an accident occurs in the scheduling process. The traditional solutions under these circumstances are mostly involved with rescheduling, which will interrupt the original task sequence in case of emergency. However, such method are not capable of satisfying the temporal constrains and the complicated evolution of events. Therefore, the study of emergency decision-making (dynamic scheduling) is imminent.

In the emergency process, we often encounter high dynamic and timeliness emergency tasks. When there is an emergency, in order to maximize the completion of the original task, the optimal decisions have to be compromised. In recent years, unmanned system can complete difficult tasks such as search and rescue~\cite{1:2013}, ~\cite{2:2016}, ~\cite{3:2016}, space underwater detection~\cite{4:2016}, surveillance and target tracking~\cite{5:2013}, ~\cite{6:2015}, transmission and communication in complex environment, and has the characteristics of autonomy, flexibility, low operation cost, good availability, more safety and efficiency, while systems modeled in multi-agents have become a research hotspot in various fields. By the recent reports and surveys, the unmanned system such as robots or UAVs have shown significant potential of reducing the cost and time needed for dynamic scheduling~\cite{7:2017}. Compared with the traditional rescue teams, the unmanned system can substitued human in the rescue task of dangerous accidents, saving considerable risk and costs.

In this paper, we mainly focus on the dynamic scheduling problems for unmanned system to solve EDUA problem in the case of time limit. One of the challenges of using unmanned systems is to coordinate their tasks.Specifically, in an emergency scenario, there is a group of task sequence. The unmanned system can solve the accident on the original task sequence without interrupting the original task sequence, and ultimately ensure the normal operation of the original task sequence and complete the rescue work, that is, the agent needs to execute the task before the deadline of the task.The main two goals are: 1) to maximize the benefits of the original task; 2) Minimize the impact of the accident. We propose a new framework called T-MADDPG), which combines transformer and MADDPG to solve EDUA problem. In the dynamic complex environment~\cite{8:2020}, multi-agent needs to complete different tasks, such as rescue, rescue resources. It can effectively rescue the sudden accident in the confined space, and make the best decision dynamically according to the accident spread. The flexible deployment of multi-agent team can effectively avoid collision and cooperate well with the team~\cite{9:2018}. The experimental results verify the ability of the Multi-Agent Reinforcement Learning in the process of solving emergencies. The new algorithm proposed in this paper is especially suitable for emergency decision-making, and can also be extended to other similar scenarios.

The main contributions of this paper can be summarized as follows.
\subsubsection{}
We propose a novel model for evaluating strong temporal-constrained emergency response tasks. We have In this model, the temporal expand and potential hazard. The advantage of our model is that it can be applied to the environment is determined while the emergency response and dispath tasks are uncertain, which are common in unmanned logistic workshops and factories. 
\subsubsection{}
Unlike the majority of forecasting models in literature, we create a general approach that is flexible to accommodate both the recessive and dominant events with MADDPG.
\subsubsection{}
We have developed three representative scenarios for simulation and evaluation. The performance of multi-agent in EDUA is tested, and the results show the superiority of the proposed Transformer-MADDPG method.

The rest of this paper is organized as follows. Section II provides some related work about the application of dynamic scheduling and the application of unmanned system in dynamic scheduling, corresponding to the content of introduction. Section III gives the definition of EDUA. In section IV, the problem of multi-agent EDUA is further described.In section V, we propose a transformer- MADDPG method, which can predict the accident situation and the trend of accident spread, and finally solve the multi-agent EDUA problem.Section VI gives the test setup and analyzes the test results. Finally, Section VII draws the conclusion and suggests the future work.

%I introduction

%II Related Work
\section{RELATED WORK}
%1%
Emergency response management (ERM) is a challenge faced around the globe. 
First responders need to respond to a variety of incidents such as fires,traffic accidents, and medical emergencies. 
They must respond quickly to incidents to minimize the risk to humanlife~\cite{II-1-1:2015,II-1-2:2014}.
Consequently, considerable attention in the last several decades has been devoted to studying emergency incidents and response.
The definition of emergent incidents in the traditional approach can be divided into two categories~\cite{II-1-3:2020}.
One way to categorize incidents is by the rate at which they occur and how they affect first responders. 
For example, some incidents happen often, and addressing them is part of day-to-day first-responder operations. 
Examples of such incidents include crimes, accidents, calls for medical services, and urban fires. 
A second category consists of comparatively less frequent incidents, which include natural calamities like earthquakes, floods, and cyclones.
The emergent incidents we study in this paper are unconventional,which have the following characteristics: they are time-sensitive, have a large number of documented cases, and spread in the environment after they occur.

%2%
In many cases, large area disasters are usually caused by small-scale emergent incidents on a small scale.
Large area disasters could be possibly be prevented if the incipient small-scale emergent incidents are detected in their early stages. 
The past decade has seen effective proposals to approach this problem with the deployment of multiple mobile sensors for monitoring emergent incidents prone areas.
and allow full map coverage.
%可删减1%
Compared to static sensor networks, robotic sensor networks offer advantages such as active sensing, large area coverage and anomaly tracking. 
%可删减1%
In 2015 David Saldana et al.~\cite{II-2-1:2015} studied the coordination and control problem of dynamic anomaly emergent incidents detection of multiple robots in the environment and proposed a distributed multi-robot dynamic anomaly detection and tracking method.
Using a probabilistic approach that integrates full-map coverage and tracking of multiple dynamic anomalies with a team of robots. In the searching phase, robots are guided toward the spots with high probability of the existence of an emergent incident.
%可删减2%
The main contribution of this work is the combination of efficient exploration under uncertain conditions, emergent incident tracking and autonomous online assignment of intelligent bodies. 
The robot explores working areas that maintain the history of the sensed area to reduce redundancy and allow full map coverage.
%可删减2%
In~\cite{II-2-2:2012}, the authors proposed an algorithm for cooperative detection of rapidly changing regions based on statistical estimation.
In~\cite{II-2-3:2020}, a comprehensive emergent incidents anomaly detection system is proposed.This system uses deep neural networks to monitor critical infrastructures such as warehouses, airports, and ports.
In~\cite{II-2-4:2018},the authors proposed an on-line monitoring frame-work for continuous real-time safety/security in learning-based control systems.
In~\cite{II-2-5:2020}, a surveillance emergent incidents detection method based on UAV-acquired video is proposed.

%3%
Mukhopadhyay A et al.~\cite{II-3-1:2020} proposed that the overall problem of emergency response to emergent incidents is actually more than just dispatching rescue personnel to the scene of an incident; specifically, emergency response can be divided into three sub-processes: (a) incident prediction, (b) resource allocation, and (c) handling computer aided dispatch to handle the emergency conditions.

%4%
Emergent incidents prediction is necessary to understand the likely demand for emergency resources in a given region. 
An important consideration when designing a prediction model is the frequent dynamic changes in the environment. 
Therefore, emergent incidents prediction methods should take into account the changes in the environment~\cite{II-4-1:2020}.
Recently, researchers have developed online models for predicting incidents, using input data streams to continuously update the learned models~\cite{II-4-2:2019}.
In~\cite{II-4-3:2017}, the authors propose a new emergent incident prediction mechanism that establishes a nonlinear mathematical procedure that provides maximum coverage of desired events.
The prediction model overcomes the problems of traditional prediction models that do not consider the priority of incident importance, spatial modeling that considers discrete regions independently, and learned models that are homogeneous by combining incident arrival time with incident severity.
Ayan Mukhopadhyay et al. in~\cite{II-4-4:2021} proposed the use of a predictive statistical model in emergency management for predicting the spatial and temporal likelihood of statistical emergent incidents occurring. The model uses a combination of synthetic resampling, non-spatial clustering, and from data-based methods to predict the spatio-temporal dynamics of incident occurrence and assign first responders across a spatial region to reduce response time.
In~\cite{II-4-5:2020}, the authors propose a general approach to emergent incident prediction that is robust to spatial variation, addressing the problem of events shifting in space.

%5%
For the scheduling processing of emergent incidents, traditional and intelligent scheduling methods are usually used. Traditional methods such as optimization methods, heuristic methods, simulation methods. Intelligent scheduling methods such as expert systems, neural network methods, intelligent search algorithms, multi-agent methods.
The most used traditional methods are heuristic scheduling algorithms and simulation methods. Intelligent scheduling methods such as expert systems, neural networks and genetic algorithms.
However, traditional methods for solving scheduling problems are usually not suitable for emergency decision-making for emergent incident(EDEI) problems, and most traditional algorithms tend to be too inflexible and slow when faced with large-scale scenarios in which the above-mentioned accidents occur~\cite{II-5-1:2008}, and end up providing suboptimal solutions that are not applicable to the problem studied in this paper.

%6%
The Multi-agent method has become one of the hot spots for research because of its speed, reliability and scalability.
Mukhopadhyay A et al. in~\cite{II-6-1:2020} proposed two partially decentralized multi-agent planning algorithms for the problem of myopic and straight-forward decision policies present in ERM systems.
In~\cite{II-5-1:2008}, a dynamically evolving scheduling mechanism is used to combine centralized and distributed policies. Global optimal scheduling is first implemented and a fast rescheduling solution is invoked in case of changes.
In~\cite{II-6-3:2012}, a mediator mechanism is used to help agents dynamically find other agents that can contribute to a given task.
Unmanned systems for problems of this type are scarce in the literature.
In 2020, Shaurya et al.~\cite{II-6-4:2020} proposed the emergency task processing in task allocation of multi robot team, and proposed a pruning heuristic method for the original task to add the emergency task. 
In 2018, Joanna Turner et al.~\cite{II-6-5:2018} proposed a distributed task rescheduling algorithm based on time constrained task allocation optimization for multi robot systems. The main challenge is to achieve the optimal allocation, maximize the completion of the original task sequence and minimize the harm of the accident.
Duan T et al.~\cite{II-6-6:2020} proposed a dynamic fault-tolerant task scheduling model (DSMFNA) to solve the problems of node failure and inability to dynamically provide capacity requirements during UAV task execution,and proposed a flexible network structure dynamic scheduling algorithm(FDSA).And a flexible network architecture dynamic scheduling algorithm (FDSA) is proposed.
However, these architectures are not suited to the problems of interest in this paper (EDEI problems)as here.

\section{PRELIMINARIES AND PROBLEM STATEMENT}

In this section, we will introduce the formal definition of the multi-agent emergency decision-making for emergent incident(EDEI) and key concepts . Then, we define the general problems of multi-agent emergency decision . 
%\textcolor{red}{ Table 1 summarizes the symbols commonly used in this paper.}

\subsection{Concept Definition}
\textbf{Definition 1}(Operation Graph). 
The graph in this paper divides the real physical space into various regions and abstracts it into a graph $\mathcal{G}=(V, E)$, where $V$ represents the set of nodes, abstracting the more important hubs or the area where the assignment needs to be performed as nodes.
The attribute $w_i$ of $V$ represents the number of assets for node $v_i$. 
The nodes are divided into three sets according to whether an incident occurs and whether the incident is completely destroyed.
The set of normal nodes $V^n$ that are not spread by the incident, the set of nodes $V^f$ that are spread by the incident and then have an incident, and the set of scrapped nodes $V^s$ that are completely destroyed by the incident.
$E$ denotes the set of edges between nodes that physically connect physical channels.
The attributes $d_{i,j}$ of $E$ represent the distance between nodes $v_i$ and $v_j$.

\textbf{Definition 2}(Asset).
The assets represent the device assets available to complete the  assignment and are denoted by $S=\{(s_1^{w_1}, ..., s_1^{w_i}), (s_2^{w_1}, ..., s_2^{w_j}), ..., (s_z^{w_1}, ..., s_z^{w_k})\}$, where $z$ denotes node $v_z$ and $w_k$ denotes the number of assets of $v_z$.
According to the damage of the node asset can determine the emergency situation of the node incident, according to the number of node asset can determine the importance of the node, so that the emergency recovery task scheduling of the agent.

\textbf{Definition 3}(Primary assignment).
The primary assignment sequence, which refers to the normal execution of the work process before an incident occurs, is denoted by $O$. Let $O=\{o_1(v_1, {et}_1, w_1), ..., o_m(v_m, {et}_m, w_m)\}$ represents the primary assignment sequence, where $v_i$ represents the node of the primary assignment $o_i$, ${et}_i$ represents the deadline of the primary assignment, and $w_i$ is the number of assets at $v_i$. The primary assignment sequence can form a single queue by the value of ${et}_i$. When performing primary assignments, a series of assets are needed. When an incident occurs, the asset at the node is destroyed, and if all the assets of a node are destroyed, the assignment of the node is judged to have failed and $s_i^j=0$ and $V^f\overset{+}{\leftarrow} \{ v_i|v_i \in O \}$.

\textbf{Definition 4}(Emergent incident). 
Emergent incident denotes a spreadable hazard that occurs at a node of the operation graph $\mathcal{G}$. The trend of hazard severity over time at the node is $f(t)$ .

\textbf{Definition 5}(Alleviate anomaly). 
In this paper, alleviate anomaly means to move assets at a node from one node to another. As in graph $\mathcal{G}$ moving assets from $v_i$ to $v_j$, i.e., $s_{k}^{w_x}\leftarrow s_{j}^{w_y}$, $s_{j}^{w_y}=0$.
The set of nodes to which the asset of a $v_i$ node can be moved is defined as $V_i^R=\{v_i^1, v_i^2, ..., v_i^k\}$, where $v_i^k\in\{V\xleftarrow[]{-}\{v_i\}\}$.
The asset at $v_i$ can be moved to any node in the set $V_i^R$.

\textbf{Definition6}(Agent). An agent refers to an unmanned vehicle that performs emergency recovery tasks, denoted as $A=\{a_1,...,a_n\}$.
The agent needs to ensure that the primary assignment sequence proceeds normally without interrupting the primary assignment sequence and with minimal impact on the primary assignment sequence.
Emergency recovery tasks are specifically classified as rescue or salvage node assets.

\subsection{Multi-agent Emergency Decision-making}
Based on the above definition, the problem of multi-agent emergency decision-making is as follows: given an undirected graph $\mathcal{G}=(V, E)$ to represent the operation graph, and the set of $n$ agent $A=\{a_1, ..., a_n\}$, a set of ordered primary assignment sequence set $O=\{o_1(v_1, {et}_1, w_1), ..., o_m(v_m, {et}_m, w_m)\}$, $m>n$, where $v_i$ represents the node of the primary assignment $o_i$, ${et}_i$ represents the deadline of the primary assignment, and $w_i$ is the number of assets at $v_i$. 
The primary assignment is an unmanned operation scenario in which the unmanned device traverses a number of nodes on the operation graph within the primary assignment deadline.
When an emergent incident occurs on graph $\mathcal{G}$, the incident spreads along the node to the surrounding nodes in the graph.
Agents need to perform emergency recovery tasks and ensure maximum completion of the primary assignment sequence.

The problem is solved by having agents complete emergency recovery tasks without conflict, so that the primary assignment sequence completes primary assignment with maximum total value and with minimum impact from incidents.

\begin{equation}
    J=\text{max}\sum  V_{alue}(O)+\text{min}\sum  V_{alue}(V^{f})
\end{equation}

Among them, $ V_{alue}(O)$ represents the most value of primary assignment completion; $V_{alue}(V^{f})$ represents the total value affected by the incident.

%IV
\section{PROBLEM DESCRIPTIONS}
In the real world, emergent incidents often affect a wide range of factors and matters, such as complex dynamic changes in the environment, time urgency, limited assets, lack of information, complex situations, and unpredictable situations.
Therefore, it is very difficult to make the right decision in a short time in an emergent incident situation.
There is no way to directly apply the traditional multi-agent emergency scheduling problem to the real world.
In the above context, we have made some simple modifications to the traditional multi-agent emergency decision making and further defined a scenario-specific problem of emergency decision-making for emergent incident(EDEI) for the type of emergency incident rescue.
The specific amendments are as follows:
\subsubsection{}
Turn the original operation graph $\mathcal{G}$ into an operation graph $\mathcal{G}$ and an incident spread graph $\mathcal{G}'$.
In real life, the operation graph $\mathcal{G}$ is abstracted from the physical space. While the incident spreads not according to the physically passable channel to determine the edge, the incident may spread along the pipeline with the circuit.
Therefore, a connection can be made to obtain the incident spread graph $\mathcal{G}'$ based on the relationship of incident spread.
Each node has materials that need to be salvaged or primary assignment that need to be completed, and the sequence of primary assignment is sequential according to the needs of the assignment.
The original map information is known in advance, and the incident will spread over time after it happens.
\subsubsection{}
Incidents spread, we are targeting incidents that can spread, such as fires.
We consider here the example of fire, where the fire situation may follow the $f(t)$ function and the spread may follow the $g(f(t))$ function.
\subsubsection{}
In this paper, an agent refers to a robot or an unmanned cart, which requires a communication mechanism between them.
Existing robots for fire rescue, such as the fully automated firefighting robot system high-pressure water gun robot and hose extension robot developed by Mitsubishi Heavy Industries in 2019~\cite{IV-1:2019}. When a fire spot is detected, the water cannon robot can automatically move into position at speeds of up to 7.2 km/h via GPS and laser sensors.
The problem is that many firefighting robots must carry a heavy water-filled hose to fire, and the robots do their jobs independently of each other, with no communication between them.
The agent needs to dynamically assign tasks to complete the emergency recovery tasks at the nodes before the incident situation $f(t)$ reaches the critical value $m$, or before the primary assignment deadline $et_i$.
\subsection{Real World Scene Modeling}
The typical incident in the emergency rescue is fire.
For example, inside an unmanned storage where a fire occurs, a realistic physical plan of the warehouse can be abstracted into an incident operation graph $\mathcal{G}=(V,E)$.
Where, $V$ is the set of assignment nodes abstracted according to different assignment areas in the warehouse, each node contains the attribute $w_i$.
When $w_{i}>0$ means that there is a asset placed at that location, its value number is $w_i$, and $w_{i}=0$ means that there is no asset at that location.
$E$ is the set of connectivity relations between assignment nodes, which are abstracted into connected edges in the graph according to the specific assignment area. 
The properties $d_{i,j}$ of $E$ denote the distance of $v_i$ from $v_j$.

In $\mathcal{G}=(V, E)$, there are a sequence of primary assignments $O=\{o_1(v_1, {et}_1, w_1), ..., o_m(v_m, {et}_m, w_m)\}$, where $v_i$ represents the node of the primary assignment $o_i$, ${et}_i$ represents the deadline of the primary assignment, and $w_i$ is the number of assets at $v_i$.
\subsection{Incident Spread}
\begin{figure}[!t]
  \centering
  \includegraphics[width=8.5cm,height=3.6cm]{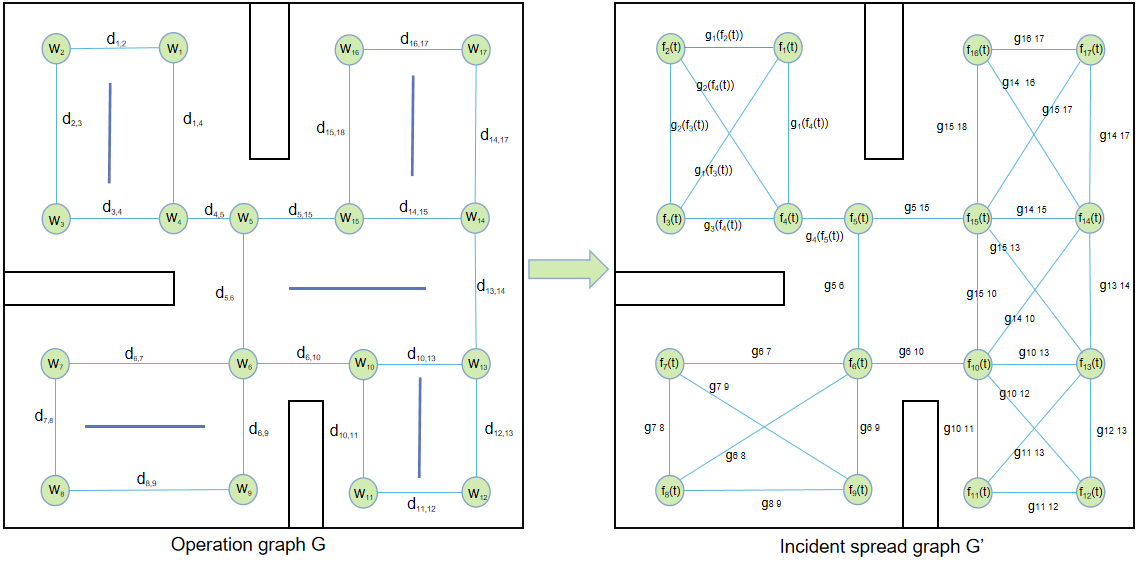}\\
  \caption{Operation graph $\mathcal{G}$ and incident spread graph $\mathcal{G}'$}
  \label{fig:IV-1G G'}
\end{figure}

Divide the real physical space into various regions to get the operation graph $\mathcal{G}$. Determine the incident spreading relationship in the physical space according to the circuit connection pipeline connection, so that the incident spreading graph $\mathcal{G}'=(V',E')$ can be generated, as in Fig.\ref{fig:IV-1G G'}.
where $|V'|=|V|$, $V'$ contains the attribute $f(t)$, which indicates the incident condition of the situation of the spatial node, specifically the severity of the incident.

%IEEE图片建议用!t，图片处于页面顶部，还可以改成ht图片处于中间

In this paper, we classify incidents into the following categories.

\subsubsection{}
Conspicuous incidents. Conspicuous incidents refer to incidents with clear incident characteristics, such as fires and shelf collapses.
\subsubsection{}
Unobtrusive incidents. Unobtrusive incidents refer to incidents that do not have clear incident characteristics, such as path congestion.

The calculation of the incident condition $f(t)$ for these two incidents has been studied extensively and accordingly.
As in ~\cite{IV-2:2006} for fires $f(t)$ was calculated using multiple low-altitude, short endurance (LASE) UAVs to explore and monitor the forest fire propagation problem, where airborne cameras to detect fires and image-based techniques were used to detect the extent of the affected area.
An unmanned aircraft system (UAS) for forest fire monitoring is presented in~\cite{IV-3:2012}, which is capable of calculating in real time the evolution of fire front patterns and other potential parameters related to fire propagation. In the paper, the area where the fire is located is divided into a rectangular grid, where the state of each cell $k$ is defined by two binary values:$F_{k,t}\in\{0,1\}$, indicating whether the cell has a fire; $Q_{k,t}\in\{0,1\}$, indicating whether the fuel in the cell is completely depleted.
In ~\cite{IV-4:2015} for the computation of $f(t)$ for shelf collapse, a cooperative exploration coordination method is proposed to detect and track multiple dynamic bounds of emergent incidents.
In [5 2020] for the calculation of $f(t)$ for path congestion, a deep neural network architecture for unsupervised anomaly detection (UAV-AdNet) is proposed to use unmanned aerial vehicles (UAV ) to monitor critical infrastructure (e.g., warehouses, airports, and ports).

In this paper, when a emergent incident occurs at a node and the incident condition $f(t)>\tau$ represents that the node is completely destroyed, the node is added to the set of $V^f$, $V^f\overset{+}{\leftarrow} \{v_i|f_i(t)>\tau \}$. Where $\tau$ is the threshold value of the severity of the emergent incident $f(t)$.

Each edge $e_{i, j}^{'}$ in $E^{'}$ connects two nodes $v_{i}^{'}$, $v_{j}^{'}$ and denotes the probability $g_j(f_i(t))$ that the incident at $v_{i}^{'}$ spreads to $v_{j}^{'}$ after an incident at $v_{i}^{'}$.
$|E|$ is not directly related to $|E^{'}|$.
After an incident, if the node is spread by the incident, the asset $w_i=0$ at this node.
If the deadline is exceeded, the primary assignment is considered failed.

After the incident spread graph is abstracted from the operation graph, the corresponding incident spread matrix $G^{*}=[g_{i,j}]\in \mathbb{R}^{N \times N}$ can be derived according to the incident condition $f_i(t)$ and incident spread probability $g_j(f_i(t))$ of the incident node, as shown in Equation~\ref{equ.matrix}.
In $G^*$, the ranks and columns represent the nodes in the incident spread graph. If there are $n$ nodes in the incident spread graph, $G^*$ is an $n*n$ matrix.
In $G^*$, the diagonal line indicates the incident condition $f_i(t)$ of the current node $v_i$, and the other corresponding positions indicate the probability of incident spread between two nodes.
For example, row $i$, column $j$, corresponding to the value $g_j(f_i(t))$, indicates that the probability of the incident at node $v_i$ spreading to node $v_j$ is $g_j(f_i(t))$, which is simplified in the matrix as $g_{i,j}$.
The current incident status of the node and the trend of incident spread can be clearly seen in the matrix.
The rows of the matrix represent the probability of spreading of the current node to the surrounding nodes, and the columns of the matrix represent the superimposed influence of the surrounding nodes on the current node.
The specific probability superimposed impact calculation formula is shown in Equation~\ref{equ.Pj}. 
Update the incident spread matrix at every time step. The parameters of the response such as the objective function are updated according to the incident spread matrix.

\begin{equation}
\label{equ.matrix}
    G^{*}=\begin{pmatrix}
        f_{1}   & g_{1,2} & \cdots & g_{1,n}\\
        g_{2,1} & f_{2}   & \cdots & g_{2,n}\\
        \vdots  & \vdots  & \ddots & \vdots\\
        g_{n,1} & g_{n,2} & \cdots & f_{n}\\
    \end{pmatrix}
\end{equation}

%$
%G^{*}=\begin{pmatrix}
%        f_{1}     & g_{1,2}   & \cdots     & g_{1,n-1}  & g_{1,n}\\
%        g_{2,1}   & f_{2}     & \cdots     & g_{2,n-1}  & g_{2,n}\\
%        \vdots    & \vdots    & \ddots     & \vdots     & \vdots\\
%        g_{n-1,1} & g_{n-1,2} & \cdots     & f_{n-1}    &g_{n-1,n}\\
%        g_{n,1}   & g_{n,2}   & \cdots     & g_{n,n-1}  & f_{n}\\
%\end{pmatrix}
%$

\begin{equation}
\label{equ.Pj}
    P_{j}=1-\prod_{i=1}^{n}(1-g_{i,j})
\end{equation}

\begin{figure}[!t]
  \centering
  \includegraphics[width=8.5cm,height=3.6cm]{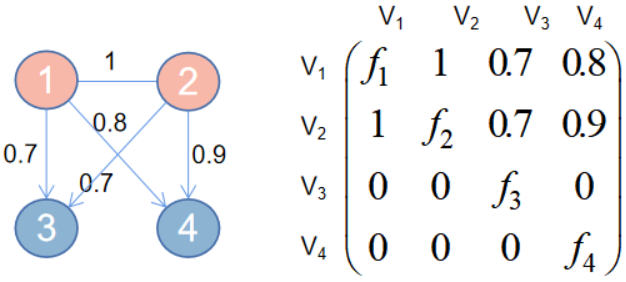}\\
  \caption{Examples of probability superposition formula and incident spread matrix}
  \label{fig:IV-2example}
\end{figure}

Example: There are 4 nodes in Figure~\ref{fig:IV-2example}, where nodes $v_1$ and $v_2$ have incidents and the probability of spreading from nodes $v_1$,$v_2$ to nodes $v_3$,$v_4$ is labeled in the figure, and the probability spreading matrix is shown on the right.

According to Figure~\ref{fig:IV-2example}, the probability of an incident at node $v_3$,$v_4$ is calculated by the probability superposition formula as follows.

$P_3=1-(1-g_{1,3})*(1-g_{2,3})=1-(1-0.7)*(1-0.7)=0.91$

$P_4=1-(1-g_{1,4})*(1-g_{2,4})=1-(1-0.8)*(1-0.9)=0.98$

In this problem, if the incidents do not spread $g_j(f_i(t))=0$, the problem can be abstracted as how to make the total execution of the primary assignment sequence and the new incoming assignment the most efficient if new assignments are continuously added to the primary assignment sequence.
If there is no primary assignment sequence, the problem can be abstracted as an ordinary rescue and relief problem, so that the damage caused by the incident is minimized.

\subsection{Agent}
Defining a set of $n$ agents $A=\{a_1, ..., a_n\}$, in which each agent can move and eliminate the incident nodes in $\mathcal{G}$ and $\mathcal{G}^{'}$. The agent starts to move at the initial position, in case of incident $f_i(t)$ reaches the critical value $m$, or completes the node assignment before the assignment deadline $et_i$. 
The emergency recovery tasks performed by the agent at the node are specifically divided into two categories: one is to perform a rescue to eliminate the incident so that $f_i(t)= 0$ or $f_i(t)<m$, and the other is to carry the assets at the node to a safe place.

The problem presented in this paper is real-time, where the emergent incident in the scenario is constantly spreading, so we assume that each agent can communicate with all other agents in the team by broadcasting messages.
At time $t$, if an incident is detected at a node, the set of incident nodes $V^f$ is updated and propagated through the communication network.
The task execution time of all the agents is inversely related to the number of agents, and more agents complete the task faster. The time required to handle the emergency recovery task of an incident node is $CT_i$, and the time taken by n agents to handle the task of the same node is $t_n=\frac{CT_i}{n}$.

\subsection{EDEI Problem}
Given an incident operation graph $\mathcal{G}$, an incident spreading graph G', and an primary assignment sequence $O$, the agent makes emergency decisions in a conflict-free situation such that the total gain is maximized when the entire primary assignment $O$ is completed.
In the above specific setup, the global optimization objective is to maximize the sum of the gains on all primary assignments, defined as follows.

\begin{equation}
    \varphi =\text{max}\sum (w_i(O)+w_i(G_{r}))
\end{equation}

Among them, $w_i(O)$ represents the total value of the completed primary assignment, and $w_i (G_{r})$ represents the total value that is not endangered in the graph $\mathcal{G}$.

%V Proposed Solutions
\section{PROPOSED SOLUTIONS}
The emergency decision algorithm used in this paper is a combination of GRU and MADDPG, where each agent makes decisions independently and shares information.
The algorithm includes three parts:
(1) dynamic perception of environmental information; (2) Predict the nodes where emergent incidents may occur in the next step; (3) The agent makes decision based on the predicted information about the nodes where incidents are likely to occur.

%（1）（2）语法有误，应该是两个名词不应该是动词，是中文有歧义导致的翻译有误
\subsection{Dynamic Perception of Environmental Information}
In the first stage of dynamic perception of environment information, agents transfer three aspects of information: (1) each agent perceives the information of changes in the surrounding environment, that is, whether there is a new emergent incident in the surrounding nodes;(2) agents share their decision information, that is, the location information of their next step; (3) Predicted node information of possible emergent incidents in the next time step.

After the agent enters the environment, it senses the incident information of the surrounding nodes, updates $V^f$ and communicates $V^f$ information with other agents.
Through the prediction mechanism, the agent predicts the nodes with possible emergent incidents and puts them into the set $\mathcal{T}_{t+1}$.
Agent communicates decision information of $t+1$ time with other agents at $t$ time.
The main steps of the algorithm are represented by pseudo code in algorithm~\ref{algorithm1}.
%算法1
\begin{algorithm}
    \caption{Dynamic Perception of Environmental Information Strategy S1.}
    \label{algorithm1}
    \begin{algorithmic}[1]  %1表示加行号
        \REQUIRE Primary assignment sequence set $O$; incident nodes set $V^f$; incident spread matrix $G^*$.
        \ENSURE $V^f_{t+1}$,$ \omega (t+1)$.
        \WHILE{$t<$ end time $T$}
            \STATE Agent perceives incident information $f_i(t)$of surrounding nodes.
            \STATE Update $V^f$, $V^{f}\overset{+}{\leftarrow}\{v_i|f_{i}(t)> 0\}$.
            \STATE Get new predict incident node set $\mathcal{T}_{t+1}$ according to algorithm\ref{algorithm2}.
            \STATE Communication $\mathcal{T}_{t+1}$ with other agents.
            \STATE Communication assignment execution list $\omega _{t}$ with other agents.
            \IF{$V^{f}(t)\neq V^{f}(t-1)$}
                \STATE Communication $V^f$ information with other agents.
            \ENDIF
            \IF{agent make new decision}
                \STATE Communication agent's own decision $u_{t+1}$with other agents.
            \ENDIF
            \STATE $t\leftarrow t+1$.
        \ENDWHILE
    \end{algorithmic}
\end{algorithm}

\subsection{Prediction Mechanism}
For the real emergent incident scene, the incident will have an impact on the surrounding area, spread rapidly to the surrounding area, and spread exponentially over time.
This will cause the environment to change rapidly all the time, and the speed of decision-making in the later stage may not keep up with the speed of incident changes.
If the agent directly explores the environment for decision-making, the decision-making efficiency will be low because of the large search space and the rapid change of the environment.
Therefore, the prediction mechanism is introduced, and the GRU is used to predict the probability of possible incidents $g_j(f_i(t))$ at the next time step node according to a large amount of common sense knowledge.
Put the nodes with $g_j(f_i(t))>0$ into the set $\mathcal{T}_{t+1}$, $\mathcal{T}_{t+1}\overset{+}{\leftarrow}\{v_i|g_j(f_i(t))> 0\}$, where $\mathcal{T}_{t+1}$ represents the 
set of nodes predicted to be incident in the next step, and share $\mathcal{T}_{t+1}$ information with other agents.

GRU can find the coupling relationship between anomalies~\cite{V-1:2021}.
When judging, GRU can not only judge the nodes that may have incidents in the next step, but also judge the compound impact on the surrounding nodes after the incident, that is, the occurrence of other types of incidents due to one incident.
For example, a fire at a node may lead to congestion of surrounding roads.
After predicting the node set $\mathcal{T}_{t+1}$ with exception in the next step, the agent selects the node with the largest $g_{i,j}$ from the $\mathcal{T}_{t+1}$ set to perform the emergency recovery task. 
Moreover, the time of emergent incident emergency recovery task is urgent, so it is unrealistic to execute the emergency recovery task in serial mode. Multi-agent executes the emergency recovery task at the same time, and establishes a comprehensive list $V^f$ of all discovered incident.

\subsubsection{Predictive Features}
In this paper, we predict incidents based on the following features.
The node that may have an incident in the next step is related to the incident feature $F_{f(t)}$ in the current environment, the asset feature $F_r$ at the node, and the node vulnerability feature $F_{\xi i}$.
\begin{itemize}
\item{incident feature $F_{f(t)}$.} Node incident status is one of the important factors affecting the incidents of other nodes. It consists of the incident state $f_i(t)$ of the node where the current incident occurs.
\item{Asset feature $F_r$.} $F_r(f_c,w_i)$ consists of two features, the asset category $f_c$ of the node and the asset quantity $w_i$ of the node.
The type and quantity of assets at the node reflect the vulnerability of the node, so it plays an important role in inferring whether an incident occurs at the node in the next step.
Some types of assets may even have a direct causal relationship with whether an incident occurs.
For example, if a node stores gasoline, the node is very vulnerable to accidents.
\item{Node vulnerability feature $F_{\xi i}$.} It reflects the degree of vulnerability of nodes without incidents.
$F_{\xi i}$ is related to the current node asset quantity $w_i$, the incident status $f_i(t)$ of the incident node and the distance $d_{i,j}$ between the current node and the incident node.
We calculate the feature according to Equation \ref{equ.5-1}.
\begin{equation}
\label{equ.5-1}
    F_{\xi i}=w_i\cdot \sum_{j=1}^{n}(f_j(t)\cdot \frac{1}{d_{i,j}})
\end{equation}
Where, $n$ represents the number of incident nodes, $w_i$ is the number of assets of the current node, $f_j(t)$ is the incident status of $v_j$ node, and $d_{i,j}$ represents the distance between $v_i$ and $v_j$.
\end{itemize}

\subsubsection{GRU Prediction Mechanism}
\begin{figure}[!t]
  \centering
  \includegraphics[width=8.5cm,height=3.6cm]{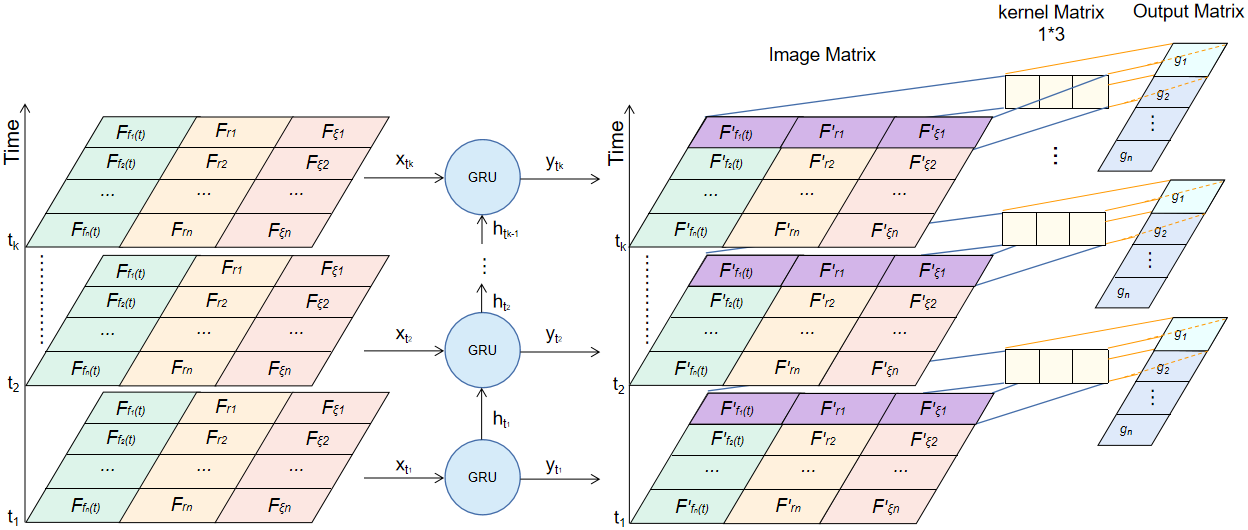}\\
  \caption{Prediction model framework}
  \label{V-1GRU}
\end{figure}

The input of GRU is constructed as an $n*4$ matrix, where $n$ is the number of nodes.
Set each row as a triple, and the elements are composed of $(F_{f(t)},F_r,F_{\xi i})$.
Output the triplet predicted in the next step $(F_{f(t)}^{'}, F_{r}^{'}, F_{\xi i}^{'})$.
The output triplet is mapped into an n*1-dimensional vector through a convolution neural network(CNN) with a convolution kernel of 1*3, which represents the incident probability $g_{i,j}$ of the next time node.
The specific prediction framework is shown in Figure \ref{V-1GRU}.

Put the nodes with possible incidents predicted by GRU into a set $\mathcal{T}_{t+1}$, 
$\mathcal{T}_{t+1}=\{<v_1, g_1>, <v_2, g_2>, ..., <v_i, g_i>\}$. 
Agents share the incident prediction set $\mathcal{T}_{t+1}$.
The main steps of GRU prediction mechanism are represented in algorithm~\ref{algorithm2}.
%算法2
\begin{algorithm}
    \caption{Prediction mechanism Strategy S2.}
    \label{algorithm2}
    \begin{algorithmic}[1]
        \REQUIRE incident feature$F_{f(t)}$, asset feature $F_r$, node vulnerability feature $F_{\xi i}$.
        \ENSURE $\mathcal{T}_{t+1}$.
        \WHILE{$t<$ end time $T$}
            \STATE $GRU(F_{f(t)},F_r,F_{\xi i})\rightarrow  \mathcal{T}_{t+k}$.
            \STATE $ \mathcal{T}_t \leftarrow \mathcal{T}_{t+1}$.
            \STATE $t \leftarrow t+1$.
        \ENDWHILE
    \end{algorithmic}
\end{algorithm}

\subsection{MDP Problem Construction}
We transform the EDEI problems into a partially observable Markov game for $N$ agents $A=\{a_1,...,a_n\}$. Define the Markov game for N agents as a set of states $S$, a set of observations $x=\{x_1,...,x_n\}$, and a set of actions $U=\{u_1,...,u_n\}$.
The state set $S$ describes the incident node set $V^f$, the primary assignment set $O$, the primary assignment completion $\omega (t)$, the location information of each agent $p_i(t)$, and the target location information of each agent $g_i(t)$.
$X$ are observation spaces for the agent, and the observation of each agent at step t is a part of the current state, $s(t)\in S$.
$U$ are action spaces for agent. For each given state $s\in S$, the agent use the policies, $\pi:S\rightarrow U_m$, to choose an action from their action spaces according to their observations corresponding to $s$.
\subsubsection{Environment State}
The state at $t$ is represented by $s(t)$, which consists of the tuple ${I(t),A(t)}$, where $I(t)$ is the set of primary assignments waiting for the service, ordered according to the relative time of incident occurrence.
$I(t)$ is a set $\{O,\omega (t),V^f_t\}$, where $O$ denotes the primary assignment sequence, $\omega (t)$ denotes the completion of $O$ at moment $t$, and $V^f_t$ denotes the set of nodes where an incident occurs at moment $t$.
$A(t)$ corresponds to the information of the agent set at step $t$, $|A(t)|=|A|$.
Each element $a_i(t) \in A(t)$ is a set of $\{p^i(t),g^i(t)\}$, where $p^i(t)$ is the current node position of agent $a_i$ and $g^i(t)$ is the target node position moved by agent $a_i$.
Then, according to the EDEI problems the environment state at step $t$, $s(t) \in S$, can be given by

\begin{subequations}
\label{state}
\begin{align}
&S(t)=\{I(t),A(t) \} \label{stateA}\\
&I(t)=\{O,\omega (t),V^f_t \} \label{stateB}\\
&|A(t)|=|A|, a_i(t) \in A(t),a_i(t)={p^i(t),g^i(t)} \label{statec}
\end{align}
\end{subequations}

\subsubsection{Action}
In our problem, the action corresponds to directing the agent to a valid node, either to process the incident or to salvage assets from the primary assignment node.
Valid nodes include nodes that have unprocessed primary assignments or nodes where incidents occur.
The action at moment $t$ is represented by $u(t)$, which consists of the tuple $\{U_O(t),U_I(t)\}$, where $U_O(t)$ denotes the processing of the primary assignment and $U_I(t)$ the processing of the incident node.
According to the current observation, each agent chooses an action from its own action space. agent's behavior at step $t$, $u(t) \in U$, can be described as follow.
\begin{equation}
\label{action}
    U(t)=\{ U_O(t),U_I(t) \}
\end{equation}

\subsubsection{Reward}
As the reward leads each agent to its optimal policy and the policy directly determines the node where the emergency recovery tasks is to be performed next time, the reward function should be designed based on the objectives of the original formulated problems,
(1) Minimize the impact of incident spread (2) Maximize the completion of the primary assignment.

In this paper, the objective function of each agent can be divided into three parts $W_{succ}$, $W_{is}$ and $W_r$, Where $W_{succ}$ represents the completion of the primary assignment, 
$W_{is}$ represents the impact of incident spread, and $W_r$ represents the remaining nodes in the graph $\mathcal{G}$.
The objective function is as follows
\begin{equation}
\label{equ.reward}
    r_{i,t}=\sum_{t=1}^{T}(W_{succ}+W_{is}+W_r)
\end{equation}

\begin{itemize}
\item{The first part shows the completion of the primary assignment, specifically refers to the rescue node assets, and is set as $W_{succ}$.} 
The completion status of initial task $O$ at each step is $\omega (t)$. 
Let the set of task nodes to be completed be $\Lambda =|O|-|\omega (t)|$. When $\Lambda > 0$, $t= T$, there are primary assignments that have not been completed. When $\Lambda = 0$ and $t \leqslant T$, it means that all primary assignments are completed.
Specifically, it can be divided into pending tasks $\Lambda$ and completed tasks $\omega (t)$, $\Lambda \cup  \omega (t)= O$.
The task to be completed $\Lambda$ is divided into general task node (i.e. no incident node) and predicted incident node $\mathcal{T}_{t+1}$.
The predicted incident node can be divided into the incident but not completely damaged node and the predicted incident node.
$W_{succ}$ is calculated as follows

\begin{equation}
\label{w_succ}
%\begin{split}
W_{succ}=
\begin{cases}
g_i\cdot f(t)\cdot w|_{v_i\in \mathcal{T}_{t+1}\bigcap \Lambda \bigcap V_f||0<f(t)<\tau || g_i=1}, (1)\\
P_i\cdot w|_{v_i\in \mathcal{T}_{t+1}\bigcap \Lambda ||f(t)=0 ||g_i > 0},(2)\\
\frac{1}{et_i}\cdot w|_{v_i\in \Lambda ||f_i(t)=0 ||g_i=0},(3) \\
w|_{v_i\in \omega _t},(4)\\
\end{cases}
%\end{split} 
\end{equation}

Where (1) represents reward for predicting incidents in incidents but not completely destroyed, (2) represents reward for predicting incidents in tasks to be completed, (3) represents reward  for primary assignments to be completed without incident, (4) represents completed primary assignment reward. 

\item{The second part represents the impact of incident spread and is set as $W_{is}$.} 
At each step $t$, there will be incident prediction set $\mathcal{T}_{t+1}$ and incident set $V^f$. Let $\Gamma =|\mathcal{T}_{t}|-|V_{f(t)}|$, when $\Gamma> 0$, it is proved that there is no incident at the node where the incident originally occurred, that is, the decision made has an effect on the spread of the incident.

There are three factors affecting the spread of incidents: the primary assignment deadline $et_i$, the node incident status $f_i(t)$, and the distance $d_j^{i}$ from the agent $a_i$ to the task node $v_j$.
According to the above three indicators, the assignments in the primary assignment sequence $O$ are sorted according to task emergencies to generate a new task set $O_{et_i}$, $O_{f(t)}$, $O_{d_j^{i}}$. 
Note that the more urgent the tasks are, the higher the ranking.
Then, the three sets are grouped in pairs as Cartesian products, which are represented by relation matrix.
The node corresponding to the lower left corner of the matrix is the most urgent position, expressed in (0, 0) coordinates.
According to the distance $d_{(0,0)}$ between each task node and (0,0) in the figure, the most serious task node can be calculated, so that the node to perform the task in the next step can be selected.
If the distance between any node and (0, 0) is equal, go to the asset quantity $Max\{v_i|_{w_i}\}$ node at the node. If the asset quantity is equal, randomly select a node for rescue.
After selecting the most serious node in the three Cartesian products, the accident spread degree is calculated by equation \ref{eqV-3}.

\begin{equation}
\label{eqV-3}
W_{is}=\left\{\begin{matrix}
\frac{1}{et_i}\cdot f(t)\cdot w_i^{O_{et_i}\times O_{f(t)}}\\ 
\frac{1}{et_i}\cdot\frac{1}{d_j^{i}}\cdot w_i^{O_{et_i}\times O_{d_i}^{v_j}}\\ 
f(t)\cdot \frac{1}{d_j^{i}}\cdot w_i^{O_{d_i}^{v_j}\times O_{f(t)}}
\end{matrix}\right.
\end{equation}

For example, if the primary assignment set $O = \{o_1, o_2, o_3, o_4\}$, it is $O_{et_i} = \{o_1, o_2, o_3, o_4\}$ from small to large according to the assignment deadline, $O_{f (t)} = \{o_2, o_4, o_3, o_1\}$ according to the incident status $f(t)$, and $O_{d_j^{i}} = \{o_3, o_1, o_2, o_4\}$ from small to large according to the distance $d_j^{i}$ between agent $a_i$ and each task node.
After that, the three sets do Cartesian product in pairs to generate the relationship matrix $O_{et_i}\times O_{f(t)}$, $O_{et_i}\times O_{d_j^i}$, $O_{d_j^i}\times O_{f(t)}$, as shown in Fig.\ref{V-3 dikaer2}. The asterisk in the figure indicates the most urgent position.
It can be seen from the calculation in (a) that node $v_2$ is the most urgent, (b) node $v_1$ is the most urgent, (c) the distance between $v_2$ and $v_3$ from node (0, 0) is the same. By comparing the size of $v_2$ node asset $w_2$ and $v_3$ node asset $w_3$, select the node with larger $w$ for rescue. If the number of assets is equal, select a node at random.
\begin{figure}[!t]
  \centering
  \includegraphics[width=8.5cm,height=3.2cm]{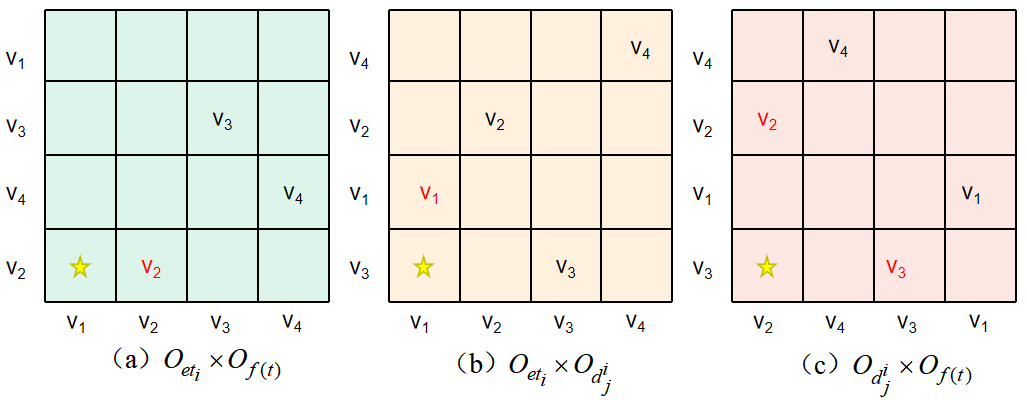}\\
  \caption{Anomalous spread affects Cartesian product (!!change describe).}
  \label{V-3 dikaer2}
\end{figure}

\item{The third part shows the remaining nodes in the graph $\mathcal{G}$, set as $W_r$.}
Specifically, it refers to the node assets that are not damaged in the graph $\mathcal{G}$. Let $V^r$ represent the set of remaining nodes in the graph $G$ that are not destroyed,$V^{r}\overset{-}{\leftarrow}\{v_i|_{et_i=0||v_i\in V_f}\}$.
The specific calculation formula of $W_r$ is as follows:
\begin{equation}
    W_r=w|_{v_i \in V^r}
\end{equation}
\end{itemize}

\subsection{P-MADDPG}
\subsubsection{MADDPG}
To solve the above Markov game for N agents, For each agent,indicated by equations \ref{state} and \ref{action}, the observation are continuous and action spaces are discrete.
Moreover, to avoid spectrum and time cost on wireless communications among different agents, because it may not be able to provide a reliable and good communication channel in the extreme environment of emergent incident, in order to ensure reliability and security, the algorithm needs to be able to complete the task well without reliable channel.

MADDPG Framework: The MADDPG framework is composed of the N agents. MADDPG adopts centralized training and decentralized execution(CTDE). Next, we take an example to explain how to centrally train the MADDPG model and execute the learned model in a decentralized way.

In the centralized offline training stage, in addition to the local observation, extra information, i.e. predicted anomaly information $\mathcal{T}_{t}$.
Namely, at step t, $\mathcal{T}_{t}$ is saved into the agent’s replay buffer with $\{V^f_t,O,\omega(t),p^i(t),g^i(t) \}$ together.
For the i-th transition of the agent,$(s_i,u_i,r_i,s’)$,we have $s_i=\{\mathcal{T}_{t},V^f_t,O,\omega(t),p^i(t),g^i(t)\}$ as shown in Fig.\ref{V-last P-MADDPG}.
When updating the parameters of the actor and the critic according to the inputted mini-batch of transitions, the actor chooses an action according to the local observation $x_i$, i.e. , $u_{i}=\mu (x_i)$, and the chosen action and $s_i$ then are valued by the critic.
Moreover, with the extra information, each agent allows to learn its state-action value function separately. Also, as aware of all other agents’ actions, the environment is stationary to each agent during the offline training stage. Thus, the biggest concern to other multi-agent RL algorithms, the dynamic environment caused by other agents’ actions, is addressed here. During the execution stage, as only local observation is required by the actor, each agent can obtain its action without aware of other agents’ information.

Considering the common objective of the formulated optimization problems, the agents should maximize the completion of initial tasks and minimize the impact of anomaly spread.
To achieve a cooperative Markov game, we assume the same immediate reward $r_i$ is returned to each agent.
The $r_i$ of each time step $t$ is calculated as follows
\begin{equation}
    r_{i,t}=\sum_{t=1}^{T}(W_{succ}+W_{ia}+W_r)
\end{equation}

\subsubsection{P-MADDPG}
\begin{figure}[!t]
  \centering
  \includegraphics[width=8.5cm,height=4.8cm]{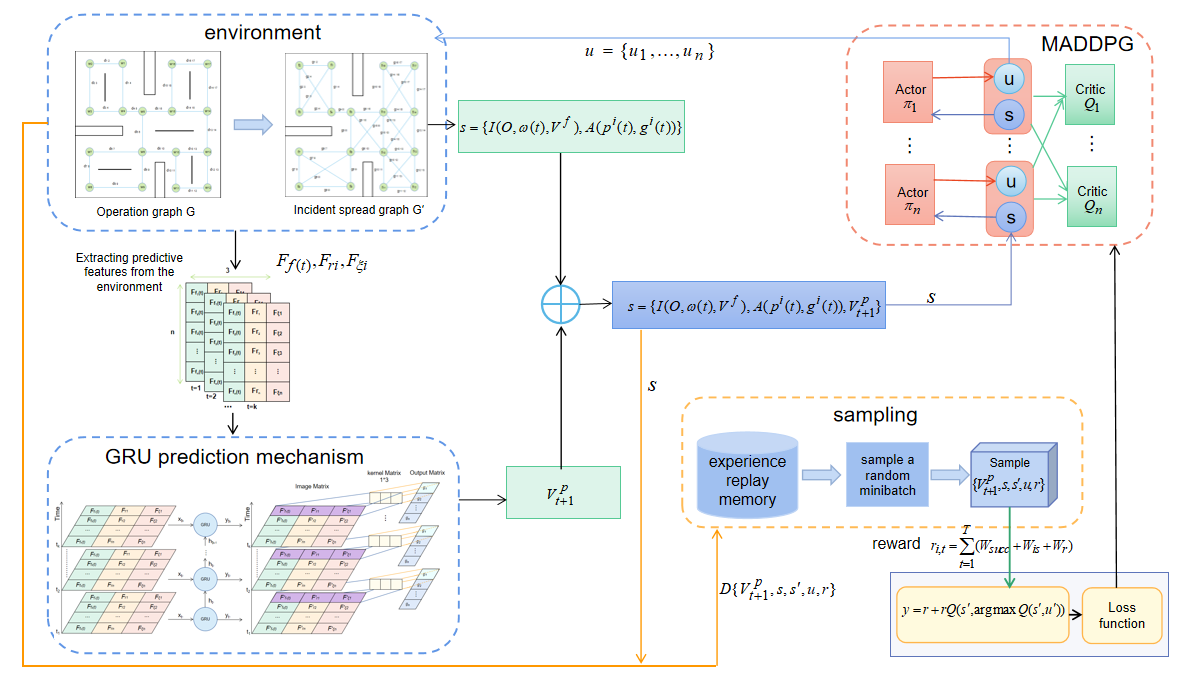}\\
  \caption{Structural framework of P-MADDPG.}
  \label{V-last P-MADDPG}
\end{figure}

When considering the decision, the node set $\mathcal{T}_{t}$ predicted by GRU that may have an incident in the next time step is taken as a part of information transmission, that is, as the information observed by agent, so as to make the decision more effective and timely.
Fig.\ref{V-last P-MADDPG} shows the structural framework of P-MADDPG.
According to the above discussion and Fig.\ref{V-last P-MADDPG}, the proposed P-MADDPG management scheme can be summarized in Algorithm~\ref{algorithm3}.
In algorithm 3, first, we perform a series of initialization (line1-4), initialize the primary assignment $O$, assignment completion $\omega (t)$, predict the incident node set $\mathcal{T}_{t}$ and incident node set $V^f$.
Then initialize a state and random action in (line6-7).
(line1-4) initialization information is shared between (Line8) agents.
In (line10-12), execute the action to update the objective function, update the observation value, store the $(s, u, R, s', V^f, \omega (t), \mathcal{T}_{t})$ information in the experience pool $D$, and (line13) predict the node set $\mathcal{T}_{t+1}$ that may have an incident in the next step.
(line14-16) update $\omega (t)$, $\mathcal{T}_{t}$, $V_f$, $\Gamma $ information, update status $s$. (line17-27) update actor network and critical network, and update target network parameters.
The specific steps are shown in algorithm 3.

%算法3
\begin{algorithm}
    \caption{P-MADDPG Strategy S3.}
    \label{algorithm3}
    \begin{algorithmic}[1]  %1表示加行号
        \REQUIRE Incident nodes set $V^f$;prediction incident node set $\mathcal{T}_{t}$;assignment execution$\omega _{t}$.
        \ENSURE Policy of agent $a_i$ $u_{t+1}$.
        \STATE Initialize  primary assignment set  $O$.
        \STATE Initialize  assignment execution$\omega _{t}$.
        \STATE Initialize  prediction incident node set $\mathcal{T}_{t}$.
        \STATE Initialize  incident node set $V^f$.
        \FOR{episode=1 to $M$}
            \STATE initialize locations of agents and initial incident nodes.
            \STATE receive initialize state $s$.
            \STATE receive $O$,$\omega_{t}$,$\mathcal{T}_{t}$,$V^f$ from algorithm 1,2.
            \STATE agents communicate the information of $O$,$\omega_{t}$,$\mathcal{T}_{t}$,$V^f$.
            \FOR{$t$=1 to max-episode-length}
                \STATE For each agent $a_i$,select action $u_{i}=\mu _{\theta _{i}}(s_{i},\mathcal{T}_i)+\mathcal{N}_{t}$.
                \STATE Execute action $u=(u_1,...,u_n)$ and new state $s’$.
                \STATE Get new reward $r_i$ from Eq.~\ref{equ.reward}.
                \STATE Get new $\mathcal{T}$ from algorithm~\ref{algorithm2}.
                \STATE Get new $\omega_{t}$,$V^f$ from algorithm~\ref{algorithm1}.
                \STATE Store $(s,u,r,s^{'},\omega _{t},V^{f},\mathcal{T}_{t})$  in replay buffer $D$.
                \STATE $\mathcal{T}\leftarrow \mathcal{T}^{'}$.
                \STATE $\omega_{t} \leftarrow \omega_{t}^{'}$.
                \STATE $V_{f} \leftarrow V_{f}^{'}$.
                \STATE $s\leftarrow s^{'}$.
                \FOR{agent $i=1$ to $N$}
                    \STATE Sample a random minibatch of $S$ samples $(s^{j},u^{j},r^{j},s^{'j},\omega _{t}^{j},V^{fj},\mathcal{T}^{j})$ from $D$.
                    \STATE Set $y^{j}=r_{i}^{j}+ \gamma \bar{Q}_{i}^{\mu^{'} }(s^{'j},\omega _{t}^{j},V^{fj},\mathcal{T}^{j},u_{1}^{'},...,u_{n}^{'})|_{u_{k}^{'}=\mu _{k}^{'}(s_{k}^{j})}$.
                    \STATE update critic by minimizing the loss $\mathcal{L}(\theta _{i})=\frac{1}{S}\sum (y^{j}-Q_{i}^{\mu }(s^{j},\omega _{t}^{j},V^{fj},\mathcal{T}^{j},u_{1}^{j},...,u_{N}^{j}))^{2}$.
                    \STATE Update actor using the sampled policy gradient:$ \triangledown _{\theta _{i}}J\approx \frac{1}{S}\sum \triangledown _{\theta _{i}}\mu  _{i}(s_{i}^{j})  \triangledown _{u _{i}}Q_{i}^{\mu }(s^{j},\omega _{t}^{j},V^{fj},\mathcal{T}^{j},u_1^{j},...,u_N^{j})|_{u_i=\mu _i(s_i^{j})}$.
                \ENDFOR
                \STATE Update target network parameters for each agent $i$: $\theta _{i}^{'}\leftarrow \tau  \theta _{i}+(1-\tau) \theta _{i}^{'}$.
            \ENDFOR
        \ENDFOR
    \end{algorithmic}
\end{algorithm}

In P-MADDPG, $D$ represents the experience replay buffer, and the elements are composed of $(s, s', u_1,...,u_n, r_1,...,r_n)$, recording all agent experiences.
In order to apply to EDEI problem, the empirical playback is improved, in which the observed value $s(t)=\{V^f(t),\mathcal{T}_{t},O,\omega(t),p^i(t),g^i(t) \}$.

\section{Experimental Results and Discussion}
This section introduces the experiments and results that test the performance of the proposed EDEI problem model and P-MADDPG algorithm.
In this paper, we select three experimental scenarios: unmanned storage, factory assembly line, civil aviation airport.
In order to further illustrate the effectiveness of P-MADDPG, it is compared with greedy algorithm and MADDPG.
Greedy algorithm is easy to implement and very efficient in most cases. MADDPG is the basic method of P-MADDPG.
Therefore, the proposed EDEI problem is evaluated by comparing the proposed P-MADDPG with greedy algorithm and MADDPG.

In this paper, we use the multi-agent system of unmanned vehicle to simulate the experiment, and conduct nine groups of experiments. See Table\ref{table1-1} for the specific experimental settings.
In the three experimental scenarios, different test conditions can be created by varying the number of tasks and agents.
In case of incident, the emergency recovery tasks to be completed by the agent include quickly rescuing the materials at the node, eliminating the incident and carrying out rescue work.
Finally, complete the primary assignment sequence as much as possible and minimize the incident impact.
The emergency recovery task is considered successful if the problem is solved, that is, the primary assignment is completed on time and on schedule and the incidents in the environment are eliminated.
The algorithm is implemented in Python and tested on a machine with Intel i7-9700k@3.6GHz CPU and 16GB RAM.
All algorithms using in this paper are coded or recoded in Python under Windows 10 operating system. All the simulations are run on a single laptop with an Intel Core i7 CPU 3.40GHz and 8GB of RAM.

% Please add the following required packages to your document preamble:
% \usepackage{multirow}
\begin{table}[]
\label{table1-1}
\caption{Experimental configuration} %title
\begin{tabular}{|c|c|c|}
\hline
Test scenario                           & Number of agents & Number of incidents \\ \hline
\multirow{3}{*}{unmanned storage}       & 2                & 4                   \\ \cline{2-3} 
                                        & 3                & 6                   \\ \cline{2-3} 
                                        & 4                & 8                   \\ \hline
\multirow{3}{*}{factory assembly line}  & 2                & 4                   \\ \cline{2-3} 
                                        & 3                & 6                   \\ \cline{2-3} 
                                        & 4                & 8                   \\ \hline
\multirow{3}{*}{civil aviation airport} & 2                & 4                   \\ \cline{2-3} 
                                        & 3                & 6                   \\ \cline{2-3} 
                                        & 4                & 8                   \\ \hline
\end{tabular}
\end{table}

\subsection{A.Unmanned Storage}
\begin{figure}[!t]
  \centering
  \includegraphics[width=8.5cm,height=2.8cm]{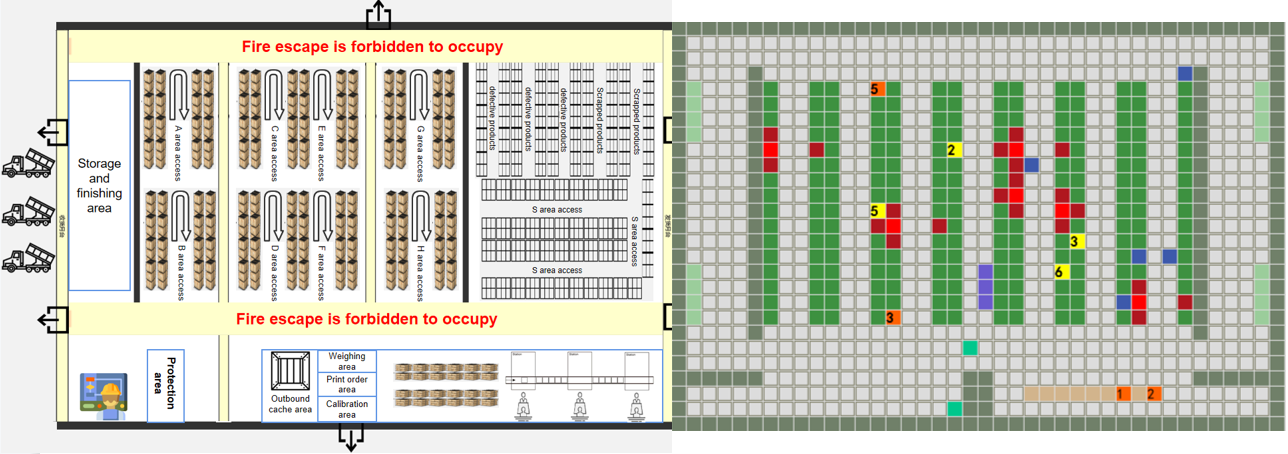}\\
  \caption{Unmanned Storage}
  \label{1-1}
\end{figure}

The experimental background selected for the first experiment is the unmanned storage, as shown in Fig.\ref{1-1}.
In the unmanned storage scenario, due to the large storage area, large number of stacking, imperfect fire safety facilities and other factors, it is easy to have safety incidents such as fire.
If the materials cannot be transferred in time, it will cause huge economic losses and affect the normal operation process such as ex warehouse and warehousing.
The working area of unmanned storage is divided into warehouse in area, warehouse out area, storage area and guarantee area.
During normal operation, the AGV transports the cargo freights from the warehouse in area to the storage area for storage, and then transports the cargo freights from the storage area to the warehouse out area for outbound.
The guarantee area is used for some support work such as charging and maintenance of AGVs.

When no exception emergent incidents, the AGV executes its own primary assignment sequence, that is, warehousing, storage and outbound tasks.
After the incident, the AGV needs to carry out cargo freight and material rescue tasks.
This is very consistent with the EDEI problem proposed in this paper, so the unmanned storage environment is selected for the experiment.

In the experimental scenario,the following experimental settings were adopted.
\subsubsection{}
The environment is defined as a 24 × 36 grid area, which represents the plan of unmanned storage. The experimental scenario is shown in Fig.\ref{1-1}.
We limit the task to 700 steps to emphasize the need for time.
\subsubsection{}
The cargo and assignment area are abstracted into nodes. There are 224 nodes in the scene, and the nodes are evenly distributed in the whole area.
\subsubsection{}
The number of resources at the node is divided according to different assignment areas. The inbound node resource $w_i=100$, the outbound node resource $w_i=100$, the support area node resource $w_i=100$, and the storage area node resource $w_i=200$.
\subsubsection{}
Agents performs the primary assignment sequence $O$. The primary assignment sequence $O$ includes warehousing and sorting assignment, storage area placement task and issue buffer assignment.
\subsubsection{}
Eight nodes are randomly selected in the scene as the initial incident nodes, that is, the red nodes in the Fig.\ref{1-1}.
\subsubsection{}
Using a team of four agents, they are in an environment with 85 primary assignment nodes. The emergency recovery tasks to be performed shall be judged according to the node incident. As in real-world scene, agents do not know the location and number of incident nodes in the environment, and achieve the conditions of local observability.
\subsubsection{}
Each time,agents can move one cell.
\subsubsection{}
At the beginning of the assignment, the starting position of the agent executing the rescue task is the warehousing area (green agent in Fig.\ref{1-1}). There is preliminary information about the location of the primary assignment.

We use greedy algorithm, MADDPG and P-MADDPG to experiment.
We evaluate the algorithms in terms of task completion rate $rate_s$, incident damage rate $rate_f$ and reward $r$.
The specific calculation formula of evaluation index is as follows:
\begin{equation*}
    rate_s=\frac{(n_{a_{1}}+...+n_{a_{i}})}{n_{O}*(n+k)}
\end{equation*}
Where, $n_{a_{i}}$ represents the number of primary assignments completed by the agent, $n_{O}$ represents the number of assignments in the primary assignment $O$, $n$ represents the number of initial agents completing the initial task $O$, $k$ represents the number of agents that change tasks after the node has an incident.

\begin{equation*}
    rate_f=\frac{n_{V_{f}}}{n_{V}*t_{max}}
\end{equation*}
Where, $n_{V_{f}}$ represents the number of nodes with incidents, $n_{V}$ represents the number of all nodes in the environment, and $t_{max}$ represents the number of steps of each episode.

\begin{figure}[!t]
  \centering
  \includegraphics[width=8.5cm,height=2.41cm]{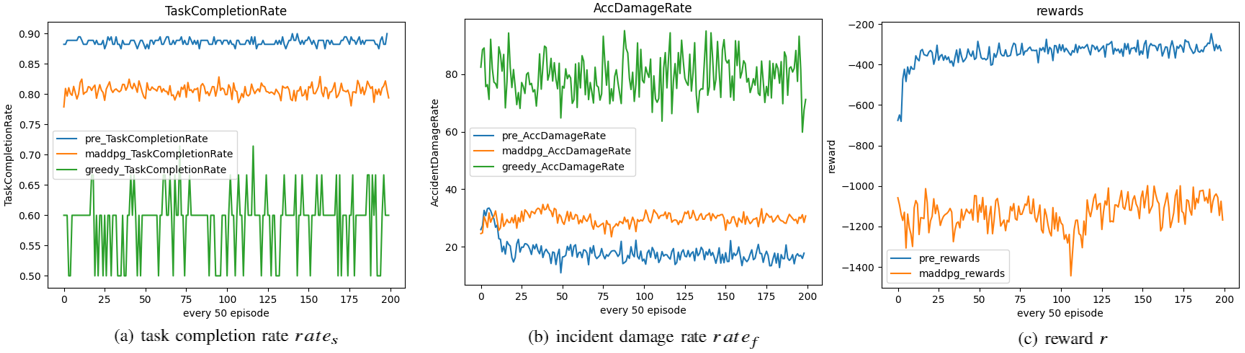}\\
  	\caption{Comparison results of three indexes of greedy algorithm, MADDPG and P-MADDPG in unmanned storage environment. (a)task completion rate $rate_s$. (b)incident damage rate $rate_f$. (c)reward $r$.}
  \label{1-3}
\end{figure}

For the experiment with 4 agents, the multi-agent adopts greedy algorithm, MADDPG and P-MADDPG respectively. After 25000 episode training, the experimental results are shown in Fig.\ref{1-3}.

In Fig.\ref{1-3}(a), among all algorithms, the task completion rate $rate_s$ of P-MADDPG is the highest, up to $90\%$, the task completion rate $rate_s$ of MADDPG is the second, up to $82.9\%$, and the task completion rate $rate_s$ of greedy algorithm is too volatile, basically maintained at about $60\%$.
In Fig.\ref{1-3}(b), the incident damage rate $rate_f$ of P-MADDPG starts to converge at 1000 episode, and the lowest incident damage rate can reach $11\%$ after convergence, and among the three algorithms, the incident damage rate $rate_f$ is the lowest.
MADDPG followed, with a minimum of $23.5\%$. The incident damage rate $rate_f$ of greedy algorithm is the worst, with a minimum of $60\%$ and a maximum of $95.1\%$.
The objective function is a unique indicator of reinforcement learning, so only the reward $r$ of P-MADDPG and MADDPG are compared in Fig.\ref{1-3}(c). In Fig.\ref{1-3}(c), compared with MADDPG, the reward of P-MADDPG will converge faster and be greatly improved after adding the prediction mechanism.

\begin{figure}[ht]
  \centering
  \includegraphics[width=8.5cm,height=2.41cm]{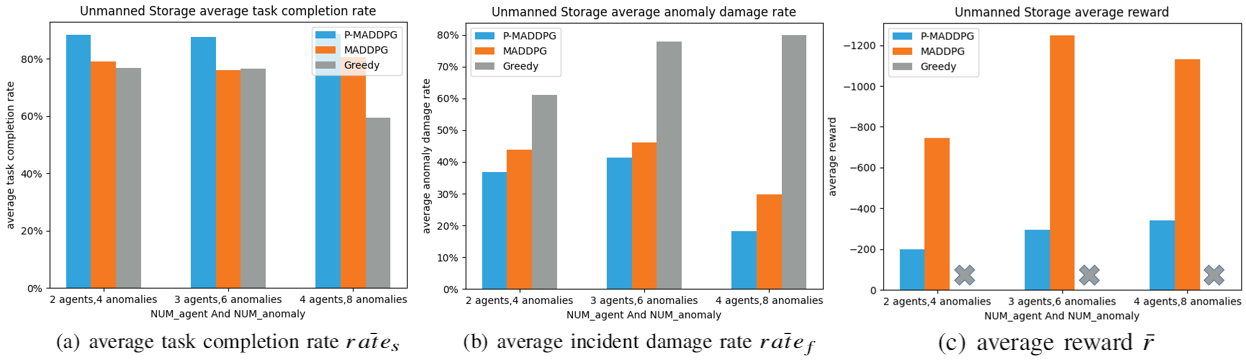}\\
  		\caption{Comparison results of P-MADDPG and other two methods in three different scenarios. (a)average task completion rate $\bar{rate_s}$. (b)average incident damage $\bar{rate_f}$. (c)average reward $\bar{r}$.}
    \label{zhu1}
\end{figure}

Note that the settings above were arbitrarily chosen and are not necessary for the algorithm to work; many other settings are possible. 
Different test conditions can be created by changing the number of initial incident nodes and agents. We designed 9 different test questions to verify the algorithm, 3 experimental configurations and 3 algorithms.
Fig.\ref{zhu1} shows the comparison results among degree, MADDPG and P-MADDPG under three experimental configurations (2 agents, 4 incidents, 3 agents, 6 incidents, 4 agents and 8 incidents).
Fig.\ref{zhu1}(a) shows the average task completion rate $\bar{rate_s}$ under three configurations.
Fig.\ref{zhu1}(b) shows the average incident damage $\bar{rate_f}$ rate under three configurations.
Fig.\ref{zhu1}(c) shows the average reward $\bar{r}$ for the three configurations.

%插入柱状图

As shown in Fig.\ref{zhu1}, each bar chart shows the average task completion rate $\bar{rate_s}$, average incident damage $\bar{rate_f}$ and average reward $\bar{r}$ when episode = 10000. 
For P-MADDPG, the task completion rate average task completion rate $\bar{rate_s}$, average incident damage $\bar{rate_f}$ and average reward $\bar{r}$ of 9 groups of experiments are the best results.

Table\ref{table1-1-1} summarizes the average of each index of all algorithms under the three configurations, and the best results of each index are highlighted in bold.

% Please add the following required packages to your document preamble:
\begin{table}[]
\label{table1-1-1}
\begin{tabular}{|c|c|c|c|c|}
\hline
Algorithm                 & \begin{tabular}[c]{@{}l@{}}Number of \\ agents and\\ incidents\end{tabular} & $\bar{rate_s}$ & \multicolumn{1}{l|}{$\bar{rate_f}$} & \multicolumn{1}{l|}{$\bar{r}$} \\ \hline
\multirow{3}{*}{Greedy}   & 2,4                            & 76.9\%         & 61.0\%                              & \textbackslash{}               \\ \cline{2-5} 
                          & 3,6                            & 76.5\%         & 77.9\%                              & \textbackslash{}               \\ \cline{2-5} 
                          & 4,8                            & 59.5\%         & 79.9\%                              & \textbackslash{}               \\ \hline
\multirow{3}{*}{MADDPG}   & 2,4                            & 79.0\%         & 43.8\%                              & -745.0                         \\ \cline{2-5} 
                          & 3,6                            & 76.1\%         & 46.1\%                              & -1247.5                        \\ \cline{2-5} 
                          & 4,8                            & 80.6\%         & 29.8\%                              & -1131.9                        \\ \hline
\multirow{3}{*}{P-MADDPG} & 2,4                            & 88.4\%         & 36.8\%                              & -197.9                         \\ \cline{2-5} 
                          & 3,6                            & 87.7\%         & 41.3\%                              & -295.4                         \\ \cline{2-5} 
                          & 4,8                            & 88.6\%         & 18.2\%                              & -341.3                         \\ \hline
\end{tabular}
\end{table}

As can be seen from table\ref{table1-1-1}, among all algorithms and experimental settings, for task completion rate $\bar{rate_s}$, when agent is 4 and incident is 8, task completion rate $\bar{rate_s}$ of P-MADDPG is the highest, reaching $88.6\%$.
For the incident damage rate $\bar{rate_f}$, when the agent is 4 and the incident is 8, the incident damage rate $\bar{rate_f}$ of P-MADDPG is the lowest, reaching $18.2\%$.
For reward $r$, when the agent is 2 and the incident is 4, the reward $r$ of P-MADDPG is the highest, reaching $-197.9$.
In conclusion, our method offers best results including the average completion rate $\bar{rate_s}$ is the largest, the average incident damage rate $\bar{rate_f}$ is the smallest, and the average reward $\bar{r}$ is the largest.

The three indexes of MADDPG are better than greedy algorithm, but worse than P-MADDPG algorithm.
When the greedy algorithm faces a complex dynamic environment and there are many incident nodes, the average anomaly damage rate $\bar{rate_f}$ of the greedy algorithm is the highest among the three algorithms, up to $76.9\%$, which does not solve the sudden incidents well.
Compared with the other three algorithms, the average completion rate $\bar{rate_s}$ of greedy algorithm is lower, and the lowest is $59.5\%$.

\subsection{B.Factory Assembly Line}
\begin{figure}[!t]
  \centering
  \includegraphics[width=8.5cm,height=2.8cm]{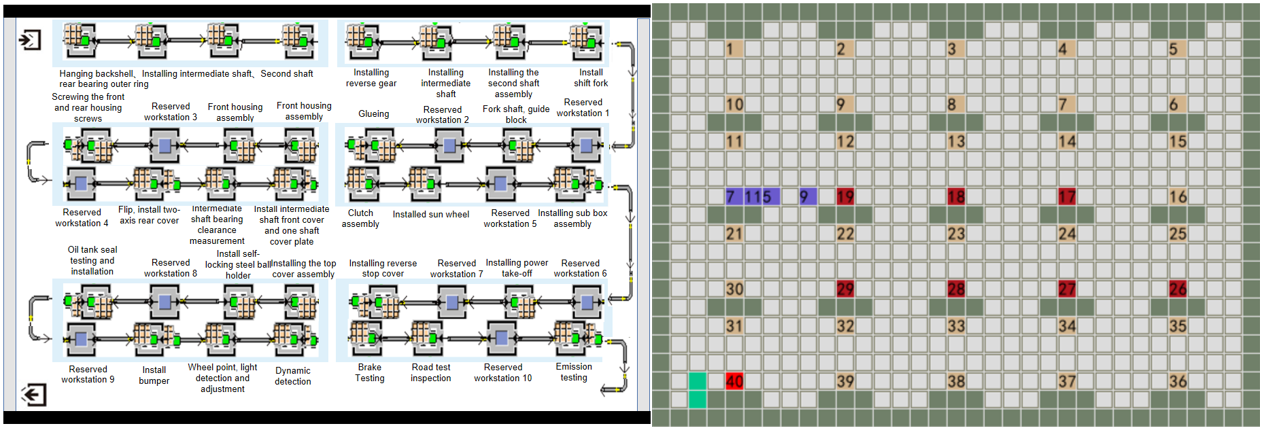}\\
  \caption{Factory Assembly Line}
  \label{2-1}
\end{figure}

The experimental background selected for the second experiment is the factory assembly line, as shown in  Fig.\ref{2-1}.
In the factory assembly line scenario, because in the actual production process, the flow shop is often affected by factors such as delayed arrival of raw materials, urgent parts insertion, change of delivery date, machine failure, parts scrapping and so on, which is easy to lead to incidents.
If the incident cannot be handled in time, the pipeline will stop running and cause huge losses.
When there is no incident, the pipeline completes the primary assignment sequence according to the primary assignment flow.
After an incident, the pipeline will reduce the task completion efficiency, and in serious cases, it will stop working.

Compared with experiment 1, the experimental settings are modified as follows. The environment is defined as a 25×32 grid area. Step is limited to 500. There are 40 nodes in the scene. Using a team of three agents. Six nodes are randomly selected in the scene as theinitial anomaly nodes. The experimental scenario is shown in Fig\ref{2-1}.

We use greedy algorithm, MADDPG and P-MADDPG to experiment.
We evaluate the algorithms in terms of though put rate $TP$, incident damage rate $rate_f$ and reward $r$.
The specific calculation formula of Though put rate $TP$ is as follows:
\begin{equation*}
    TP=\frac{n_{a_{1}}+...+n_{a_{i}}}{(n_{O}+k-1)*\bigtriangleup t}
\end{equation*}
Where, $n_{a_{i}}$ represents the number of primary assignments completed by the agent, $n_{O}$ represents the number of tasks in the primary assignment $O$, $k$ is the number of flow line segments, $\bigtriangleup t$ is the time required to perform an initial task.

\begin{figure}[ht]
  \centering
  \includegraphics[width=8.5cm,height=2.41cm]{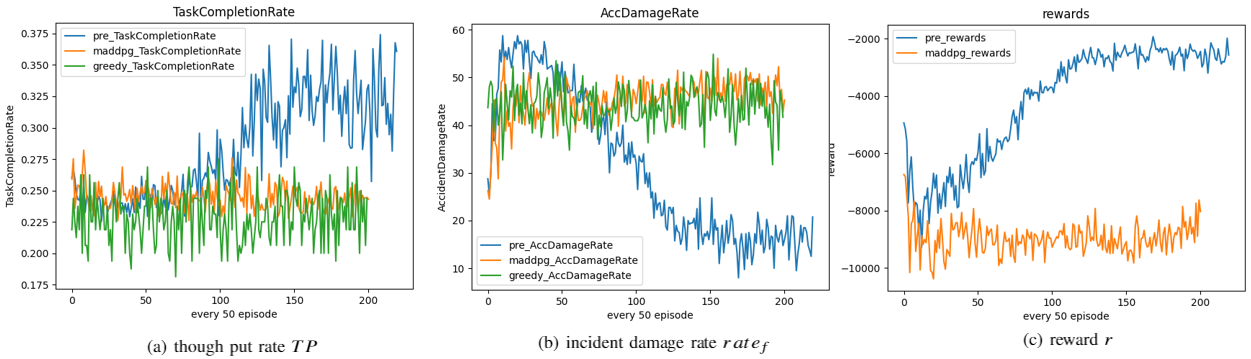}\\
  	\caption{Comparison results of three indexes of greedy algorithm, MADDPG and P-MADDPG in factory assembly line. (a)though put rate $TP$. (b)incident damage rate $rate_f$. (c)reward $r$.}
  \label{1-2-3}
\end{figure}

For the experiment with 3 agents, the multi-agent adopts greedy algorithm, MADDPG and P-MADDPG respectively. After 25000 episode training, the experimental results are shown in Fig.\ref{1-2-3}.

In Fig.\ref{1-2-3}(a), the through put rate $TP$ of P-MADDPG is the highest. It starts to converge at 5000 episode, and the maximum throughput can reach $38\%$ after convergence.
The through put rate $TP$ of MADDPG and greedy algorithm is too volatile.
MADDPG is better than greedy algorithm, maintaining at $23\%$-$28\%$.
The worst performance of greedy algorithm is maintained at $18\%$-$27\%$.
In Fig.\ref{1-2-3}(b), the incident damage rate $rate_f$ of P-MADDPG starts to converge at 7500 episode. After convergence, the incident damage rate can reach as low as $8\%$, and among the three algorithms, the incident damage rate $rate_f$is the lowest.
The incident damage rate of MADDPG and greedy algorithm is too volatile, which is basically maintained between $35\%$-$50\%$.
In Fig.\ref{1-2-3}(c), compared with MADDPG, the reward of P-MADDPG will converge faster and be greatly improved after adding the prediction mechanism.

\begin{figure}[ht]
  \centering
  \includegraphics[width=8.5cm,height=2.41cm]{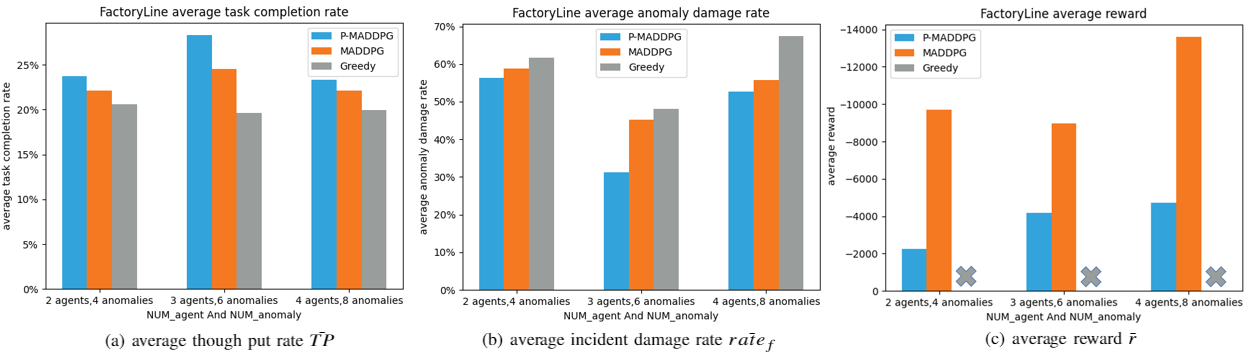}\\
  		\caption{Comparison results of P-MADDPG and other two methods in three different scenarios. (a)average though put rate $\bar{TP}$. (b)average incident damage $\bar{rate_f}$. (c)average reward $\bar{r}$.}
    \label{zhu2}
\end{figure}

Nine different test problems are designed to verify the performance of the algorithm in the factory assembly line. The experimental results are shown in Fig.\ref{zhu2}. 
Fig.\ref{zhu2} shows the comparison results among degree, MADDPG and P-MADDPG under three experimental configurations.
Fig.\ref{zhu2}(a) shows the average throughput $\bar{TP}$.
Fig.\ref{zhu2}(b) shows the average incident damage rate $\bar{rate_f}$.
Fig.\ref{zhu2}(c) shows the average reward $\bar{r}$.

%插入柱状图2

Table\ref{table1-2-1}  summarizes  the  average  of  each  index  of  all algorithms under the three configurations, and the best results of each index are highlighted in bold.

\begin{table}[]
\label{table1-2-1}
\caption{Experimental configuration} %title
\begin{tabular}{|c|c|c|c|c|}
\hline
Algorithm                 & \begin{tabular}[c]{@{}l@{}}Number of \\ agents and\\ incidents\end{tabular} & $\bar{TP}$        & $\bar{rate_f}$  & $\bar{r}$        \\ \hline
\multirow{3}{*}{Greedy}                          & 2,4                                                                         & 20.7\%            & 61.7\%          & \textbackslash{} \\ \cline{2-5} 
         & 3,6                                                                         & 19.6\%            & 48.0\%          & \textbackslash{} \\ \cline{2-5} 
                          & 4,8                                                                         & 19.9\%            & 67.4\%          & \textbackslash{} \\ \hline
\multirow{3}{*}{MADDPG}                          & 2,4                                                                         & 22.1\%            & 58.9\%          & -9718.2          \\ \cline{2-5} 
         & 3,6                                                                         & 24.5\%            & 45.3\%          & -8956.8          \\ \cline{2-5} 
                          & 4,8                                                                         & 22.1\%            & 55.8\%          & -13593.1         \\ \hline
\multirow{3}{*}{P-MADDPG}                          & 2,4                                                                         & 23.7\%            & 56.4\%          & \textbf{-2240.6} \\ \cline{2-5} 
       & 3,6                                                                         & \textbf{28.3\%}   & \textbf{31.2\%} & -4182.3          \\ \cline{2-5} 
                          & 4,8                                                                         & 23.3\%            & 52.6\%          & -4720.9          \\ \hline
\end{tabular}
\end{table}

As can be seen from Fig.\ref{zhu2}, for P-MADDPG, the average throughput $\bar{TP}$, the average incident damage rate $\bar{rate_f}$ and the average reward $\bar{r}$ of the three groups of experiments are the best results.
As can be seen from table\ref{table1-2-1}, among all algorithms and experimental settings, our method P-MADDPG provides the best results for the four experimental indicators.
The maximum $\bar{TP}$ is $28.3\%$ (3 agents, 6 anomalies), the minimum $\bar{rate_f}$ is $31.2\%$ (3 agents, 6 anomalies), and the maximum $\bar{r}$ is $-2240.6$ (2 agents, 4 anomalies).

\subsection{Civil Aviation Airport}
\begin{figure}[!t]
  \centering
  \includegraphics[width=8.5cm,height=2.8cm]{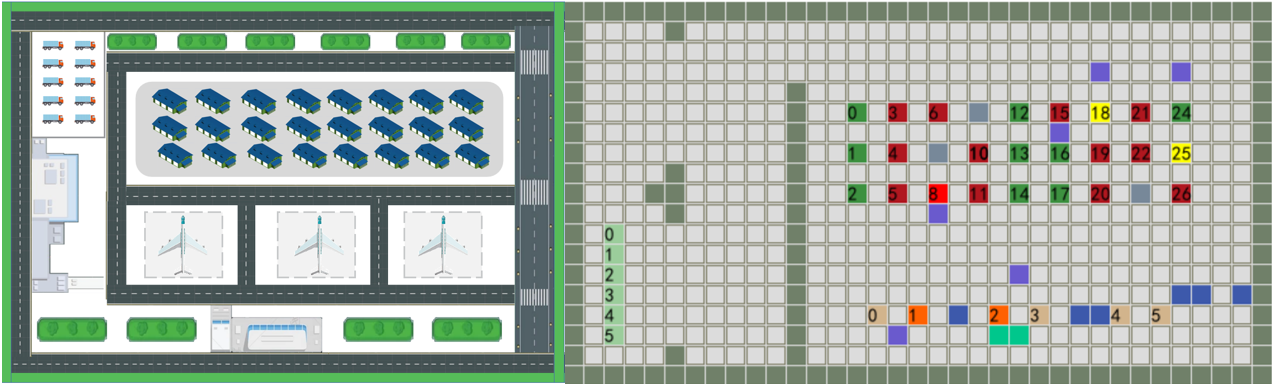}\\
  \caption{Civil Aviation Airport}
  \label{3-1}
\end{figure}

The experimental background selected for the third experiment is the baggage transportation of unmanned logistics trailers in civil airports based on Hong Kong International Airport and Hunan Airport, as shown in Fig.\ref{3-1}. The transportation and transfer of baggage and goods is an important operation content of airport logistics. Efficient logistics operation ensures the smooth and orderly connection of flights and helps to improve passenger satisfaction. In recent years, with the development of autopilot technology, landing applications have been welcomed in many scenarios, and airport logistics is one of the typical scenarios. In the actual transportation process, the unmanned logistics trailer is vulnerable to major activities, emergencies, equipment failures and other factors. If the incident is not handled in time, it will lead to baggage accumulation, reduce the supply capacity of airport resources, and can not meet the regular or sudden increase of baggage transportation demand. In serious cases, it may lead to the delay of aircraft flights and cause huge losses. The use of unmanned vehicles for emergency rescue can overcome many disadvantages brought by "human factors", can be flexibly applied to various complex airport environments and bad weather, meet the needs of all-weather and unmanned, greatly reduce labor costs and improve rescue efficiency.

In case of no incident, the goods are received in the bulk cargo receiving area by zones, the conventional goods are stacked on the pallet and enter the security inspection line. After automatic information collection, the goods information is transmitted to the security inspection machine to judge whether there is any harm. After passing the security inspection, the goods are sent to the storage area by unmanned logistics Trailer after entering the cargo station. When the board needs to be punched, the freight management system gives a signal, The unmanned logistics trailer will send the bulk cargo to be punched to the corresponding punching position, and the punching operation will be carried out manually. After the punching is completed, call the unmanned logistics trailer to send the baggage to the aircraft baggage haulage warehouse. After the incident, the unmanned car shall carry out rescue and baggage handling tasks.

Compared with experiment 1, the experimental settings are modified as follows. The environment is defined  as a 13×21 grid area. Step is limited to 180. There are 31 nodes in the scene. Using a team of four agents. Eight nodes are randomly selected in the scene as the initial incident nodes. The experimental scenario is shown in Fig\ref{3-1}.

We use greedy algorithm, MADDPG and P-MADDPG to experiment. We evaluate the  algorithms in terms of transport efficiency $TE$, inventory carry rate $IT$, incident damage rate $rate_f$and reward $r$.
The specific calculation formulas of transport efficiency $TE$ and inventory carry rate $IT$ are as follows:

\begin{equation*}
    TE=\frac{(n_{a_{1}}+...+n_{a_{i}})}{n_{O}*(n+k)*2}
\end{equation*}
\begin{equation*}
    IT=\frac{(n_{a_{1}}+...+n_{a_{i}})}{n_a*2}
\end{equation*}

Where Where, $n_{a_{i}}$ represents the number of bags transported by agent $a_i$, and $n_a$ represents the number of checked baggage.

\begin{figure}[!t]
  \centering
  \includegraphics[width=8.5cm,height=6.5cm]{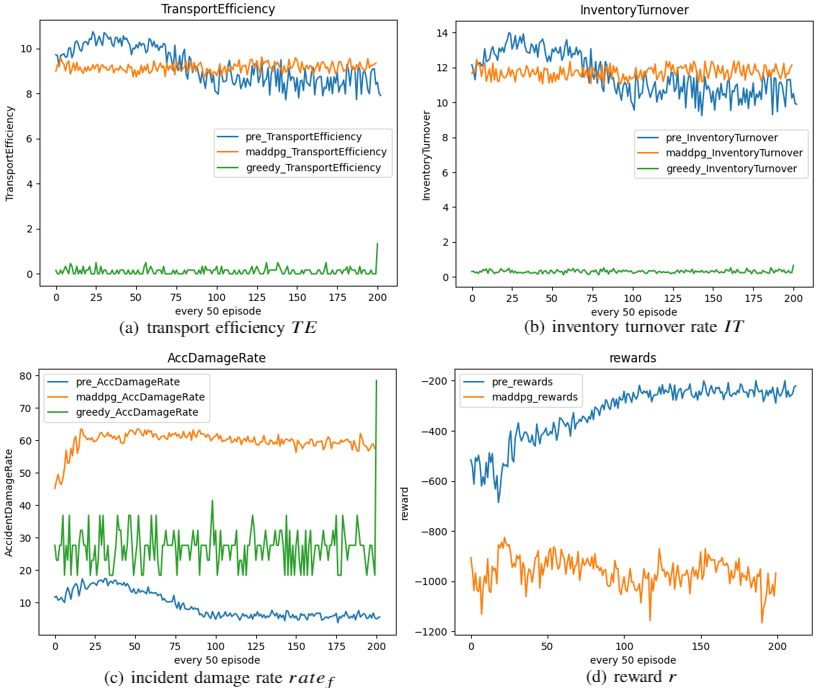}\\
  	\caption{Comparison results of three indexes of greedy algorithm, MADDPG and P-MADDPG in civil aviation airport. (a)transport efficiency $TE$. (b)inventory carry rate $IT$. (c)reward $r$.}
	\label{1-3-3}
\end{figure}

For the experiment with 3 agents, the multi-agent adopts greedy algorithm, MADDPG and P-MADDPG respectively. After 25000 episode training, the experimental results are shown in Fig.\ref{1-3-3}.

In Fig.\ref{1-3-3}(a), the transport efficiency $TE$ of P-MADDPG and MADDPG is basically the same. After P-MADDPG converges, the transport efficiency $TE$ is slightly lower than that of MADDPG. The transport efficiency $TE$ of greedy algorithm is the lowest, between 0-$5\%$. In Fig.\ref{1-3-3}(b),...
In Fig.\ref{1-3-3}(c), the incident damage rate $rate_f$ of P-MADDPG starts to converge at 5000 episode. After convergence, the incident damage rate $tare_f$ can reach as low as $5\%$, and among the three algorithms, the incident damage rate $rate_f$ is the lowest. The incident damage rate $rate_f$ of MADDPG is the worst, reaching about $60\%$ after convergence. The incident damage rate of greedy algorithm is between MADDPG and P-MADDPG, which is basically maintained between $20\%$-$40\%$.
In Fig.\ref{1-3-3}(d), compared with MADDPG, the reward of P-MADDPG will converge faster and be greatly improved after adding the prediction mechanism.

\begin{figure}[!t]
  \centering
  \includegraphics[width=8.5cm,height=6.5cm]{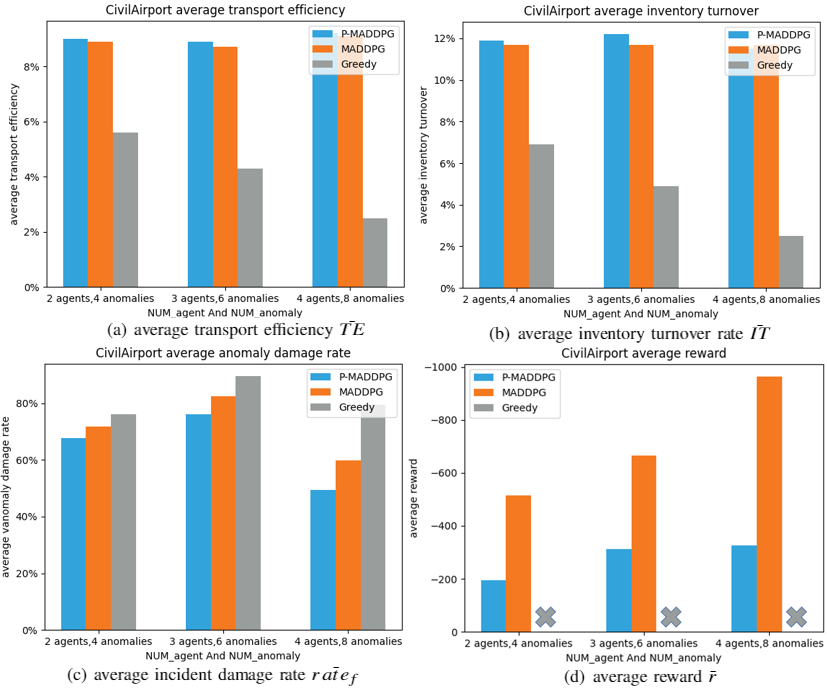}\\
  		\caption{Comparison results of P-MADDPG and other two methods in three different scenarios. (a)average transport efficiency $\bar{TE}$. (b)average inventory turnover rate $\bar{IT}$. (c)average incident damage rate $\bar{rate_f}$. (d)average reward $\bar{r}$.}
	\label{zhu3}
\end{figure}

Nine  different  test  problems  are  designed  to  verify  the performance  of  the  algorithm  in  the  civil aviation airport. The experimental results are shown in Fig.\ref{zhu3}.
Fig.\ref{zhu3} shows the comparison results among degree, MADDPG and P-MADDPG under three experimental configurations.
Fig.\ref{zhu3}(a) shows the average transport efficiency $\bar{TE}$.
Fig.\ref{zhu3}(b) shows the average inventory turnover $\bar{IT}$.
Fig.\ref{zhu3}(c) shows the average incident damage rate $\bar{rate_f}$.
Fig.\ref{zhu3}(d) shows the average reward $\bar{r}$.

%插入柱状图

Table\ref{table1-3-1}  summarizes  the  average  of  each  index  of  all algorithms under the three configurations, and the best results of each index are highlighted in bold.

\begin{table}[]
\label{table1-3-1}
\begin{tabular}{|c|c|c|l|c|c|}
\hline
Algorithm                 & \begin{tabular}[c]{@{}c@{}}Number of \\ agents and\\ incidents\end{tabular} & $\bar{TE}$     & $\bar{IT}$      & $\bar{rate_f}$  & $\bar{r}$                \\ \hline
\multirow{3}{*}{Greedy}                          & 2,4                                                                         & 6.9\%          & 5.6\%           & 76.1\%          & \textbackslash{} \\ \cline{2-6} 
         & 3,6                                                                         & 4.9\%          & 4.3\%           & 89.5\%          & \textbackslash{} \\ \cline{2-6} 
                          & 4,8                                                                         & 2.5\%          & 2.5\%           & 79.3\%          & \textbackslash{} \\ \hline
\multirow{3}{*}{MADDPG}                          & 2,4                                                                         & 8.9\%          & 12.5\%          & 71.8\%          & -515.3           \\ \cline{2-6} 
         & 3,6                                                                         & 8.7\%          & 11.7\%          & 82.5\%          & -663.9           \\ \cline{2-6} 
                          & 4,8                                                                         & 9.1\%          & 12.5\%          & 59.8\%          & -962.6           \\ \hline
\multirow{3}{*}{P-MADDPG}                          & 2,4                                                                         & 9.0\%          & \textbf{12.7\%} & 67.8\%          & \textbf{-195.7}  \\ \cline{2-6} 
       & 3,6                                                                         & 8.9\%          & 12.2\%          & 76.1\%          & -311.8           \\ \cline{2-6} 
                          & 4,8                                                                         & \textbf{9.2\%} & 11.5\%          & \textbf{49.4\%} & -327.0           \\ \hline
\end{tabular}
\end{table}

As can be seen from Fig.\ref{zhu3}, for P-MADDPG, the average transport efficiency $\bar{TE}$, the average inventory turnover $\bar{IT}$, the average incident damage rate $\bar{rate_f}$ and the average reward $\bar{r}$ of the three groups of experiments are the best results.
As can be seen from table\ref{table1-3-1}, among all algorithms and experimental settings, our method P-MADDPG provides the best results for the four experimental indicators.
The maximum $\bar{TE}$ is $9.2\%$ (4 agents, 8 incidents), the maximum $\bar{IT}$ is $12.7\%$ (2 agnts, 4 incidents), the minimum $\bar{rate_f}$ is $49.4\%$ (4 agents, 8 incidents), and the maximum $\bar{r}$ is $-195.7$ (2 agents, 4 incidents).

\section{CONCLUSION}
In this paper, we model the emergency decision-making problem for emergent incidents and propose a simple P-MADDPG algorithm combining GRU and MADDPG to solve the emergency decision-making problem for emergent incidents.
Compared with the traditional MARL method, this method uses GRU for incident prediction, and the prediction results are used as the input of MADDPG to obtain a better learning efficiency.
In the experiments, three different experimental environments were used to generate training samples, and comparative experiments were conducted with the greedy algorithm, MADDPG and P-MADDPG to test the performance of the decision-making methods under different environments, and the experimental results showed the effectiveness of the P-MADDPG algorithm.

%{\appendices
%\section*{Proof of the First Zonklar Equation}
%Appendix one text goes here.
% You can choose not to have a title for an appendix if you want by leaving the argument blank
%\section*{Proof of the Second Zonklar Equation}
%Appendix two text goes here.}

\bibliographystyle{IEEEtran}
\bibliography{ref}

\newpage

\end{document}